\documentclass[journal]{IEEEtran}

\ifCLASSINFOpdf
\else
\fi

\usepackage{times,amsmath,epsfig,subfigure,graphicx,multirow,algorithm,algorithmic,bm}
\usepackage{enumerate,paralist}
\usepackage{cite,amsfonts,dsfont,amsthm}
\usepackage[hyphens]{url}

\usepackage[footnotesize]{caption}
\begin{document}
%
\title{A Perceptually Weighted Rank Correlation Indicator for Objective Image Quality Assessment}

\author{Qingbo~Wu,~Hongliang~Li,~Fanman~Meng,~and~King~N.~Ngan
\thanks{Q.~Wu, H.~Li, and F.~Meng are with the School
of Electronic Engineering, University of Electronic Science and Technology of China, Chengdu
611731, China (e-mail: qbwu@uestc.edu.cn; hlli@uestc.edu.cn; fmmeng@uestc.edu.cn).}
\thanks{K.~N.~Ngan is with the Department of Electronic Engineering, Chinese
University of Hong Kong, ShaTin, Hong Kong, and the School of Electronic
Engineering, University of Electronic Science and Technology of China,
Chengdu 611731, China (e-mail: knngan@ee.cuhk.edu.hk).}
}


\maketitle

\begin{abstract}
In the field of objective image quality assessment (IQA), the Spearman's $\rho$ and Kendall's $\tau$ are two most popular rank correlation indicators, which straightforwardly assign uniform weight to all quality levels and assume each pair of images are sortable. They are successful for measuring the average accuracy of an IQA metric in ranking multiple processed images. However, two important perceptual properties are ignored by them as well. Firstly, the sorting accuracy (\textit{SA}) of high quality images are usually more important than the poor quality ones in many real world applications, where only the top-ranked images would be pushed to the users. Secondly, due to the subjective uncertainty in making judgement, two perceptually similar images are usually hardly sortable, whose ranks do not contribute to the evaluation of an IQA metric. To more accurately compare different IQA algorithms, we explore a perceptually weighted rank correlation indicator in this paper, which rewards the capability of correctly ranking high quality images, and suppresses the attention towards insensitive rank mistakes. More specifically, we focus on activating `valid' pairwise comparison towards image quality, whose difference exceeds a given sensory threshold (\textit{ST}). Meanwhile, each image pair is assigned an unique weight, which is determined by both the quality level and rank deviation. By modifying the perception threshold, we can illustrate the sorting accuracy with a more sophisticated \textit{SA-ST} curve, rather than a single rank correlation coefficient. The proposed indicator offers a new insight for interpreting visual perception behaviors. Furthermore, the applicability of our indicator is validated in recommending robust IQA metrics for both the degraded and enhanced image data.
\end{abstract}


%
\IEEEpeerreviewmaketitle

\section{Introduction}

In most image processing applications, human eyes are the ultimate receiver of the visual signal, where the subjective test could offer ideal evaluations for various process results. However, confronting the massive image data generated in our daily life, it has become impossible to manually annotate the image quality via substantial human subjects, which is usually time consuming, troublesome and expensive \cite{QoE_assessment}. As an alternative, it is more economical and practical to count on well-designed objective image quality assessment (IQA) metrics.

Given multiple processed or contaminated images, a main target of objective IQA is to correctly sort their quality levels via a computational model, which could approach the rank-order produced by the subjective judgement. A reliable objective IQA model not only serves as an image quality monitor, but also plays important roles in optimizing various quality driven applications, such as, image/video coding \cite{ssim_coding}, image fusion \cite{image_fusion}, contrast enhancement \cite{contrast_enhancement}, and so on. With the boom of perceptually friendly image/video processing systems \cite{IQA_application,QoE_communication,big_data_computing}, recent decades have witnessed the growing interests in the development of IQA algorithms. A large number of subject rated databases (such as, LIVE II \cite{LIVE}, TID2013 \cite{TID2013}, CSIQ \cite{CSIQ}, ChallengeDB \cite{ChallengeDB}) and objective metrics (including the full-reference \cite{FR_IQA}, reduced-reference \cite{RR_IQA}, and no-reference models \cite{RR_NR_IQA,R3}) are successively proposed to interpret the human perception of image quality from different perspectives. Thanks to the efforts of these researches, many exciting findings and technologies are verified efficient in modeling the perceptual image quality, such as, the perceptual visibility threshold \cite{JND_validity,JND}, the visual attention psychology \cite{visual_attention,saliency_induced}, the structural similarity measurement \cite{SSIM,IWSSIM,CWSSIM}, the natural scene statistics \cite{NSS_VIF,NSS_contrast}, the semantic obviousness \cite{semantic_obviousness}, and so on. Such promising progress offers us a wide variety of models to address the IQA problem. But meanwhile, it also becomes a sweet nuisance to fairly evaluate and compare them, which is crucial to guide the choice of various quality driven applications in the real world.

At present, the most common evaluation indicators for IQA are two classic rank correlation statistics, i.e., the Spearman's $\rho$ and Kendall's $\tau$ \cite{rank_correlation}. It is not surprising because the mean opinion score (MOS) collected from human subjects is typically represented by an ordinal variate in terms of rating scale \cite{ITU_rec,rate_scale}. Computing the rank correlation between IQA prediction and MOS is a straightforward way to measure its capability of sorting out perceptually preferred image. However, there is little research discussing the suitability of directly applying Spearman's and Kendall's statistics to the IQA task, even they are successful in many other fields, such as, medicine \cite{biostatistics}, biology \cite{declining_amphibian}, neuroscience \cite{Making_memories}, and so on.

\begin{figure*}[t]
\centering
  \includegraphics[width=.9\linewidth]{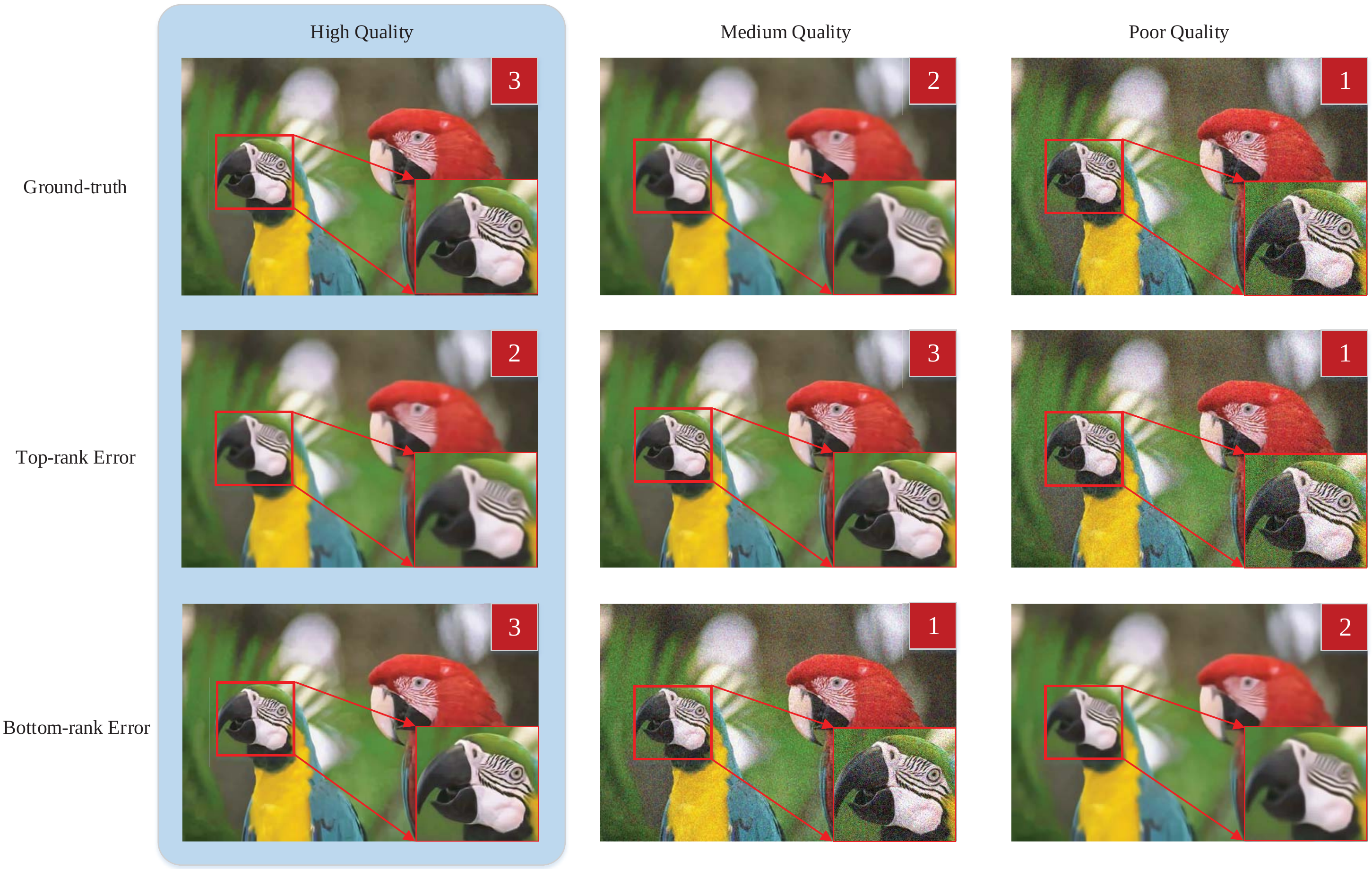}\\
  \caption{The subjective comparison between two rank results for the denoised images. The first row gives the ground-truth rank in terms of image quality, where the rank-orders are labeled in the top right corner of each image. The second and third rows show the rank results whose sorting errors occur in the top-rank and bottom-rank, respectively. Both the second and third rows share the same SRCC and KRCC, which are 0.500 and 0.333, respectively. The recommended denoising results in terms of different ranks are highlighted by the light blue window.}
  \label{fig_rank_compare}
\end{figure*}

\begin{figure}
  \centering
  \subfigure[Compressed \textit{house.bmp} (DMOS = 13.72)]{
  \includegraphics[width=.85\linewidth]{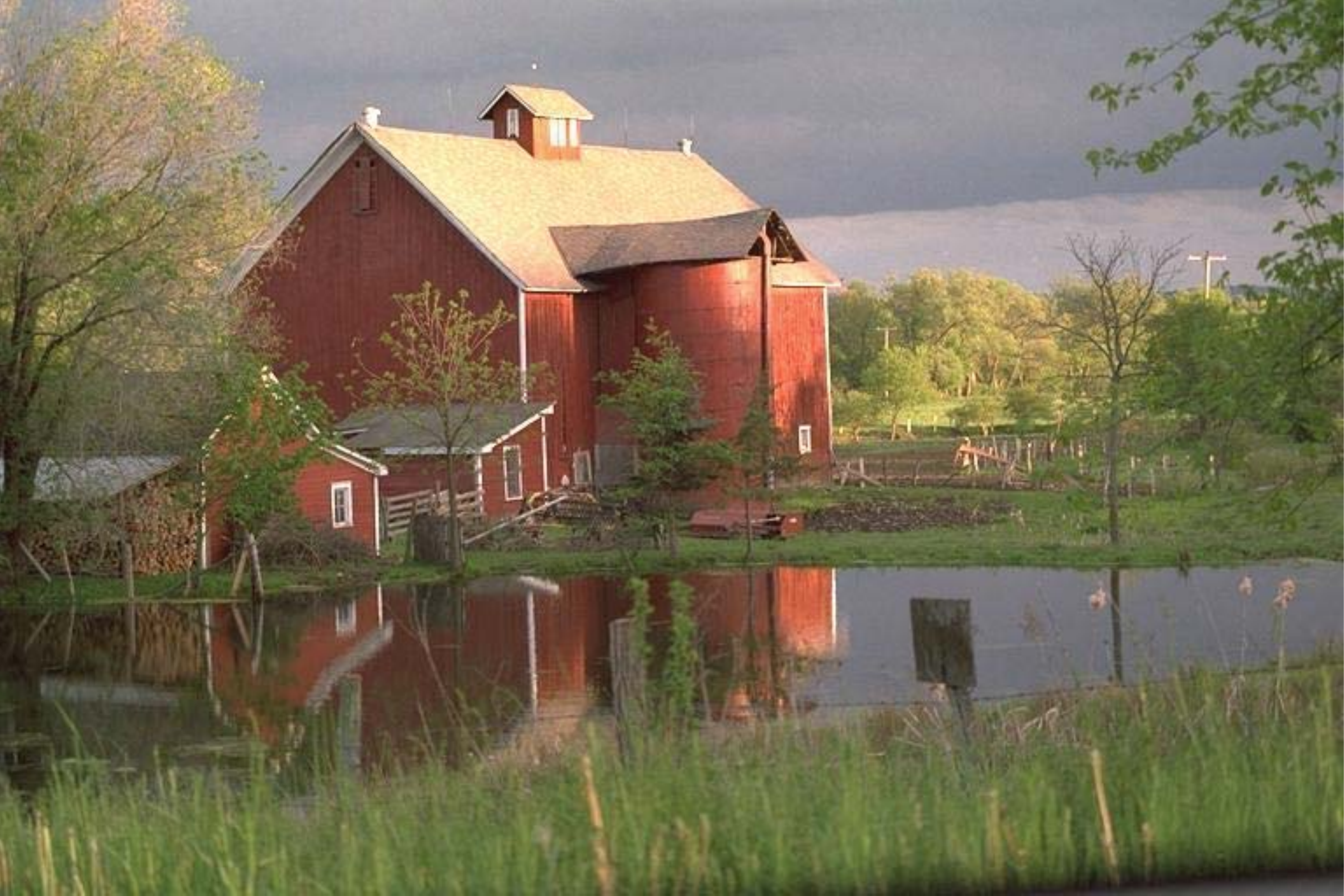}}\\
  \subfigure[Compressed \textit{buildings.bmp} (DMOS = 13.70)]{
  \includegraphics[width=.85\linewidth]{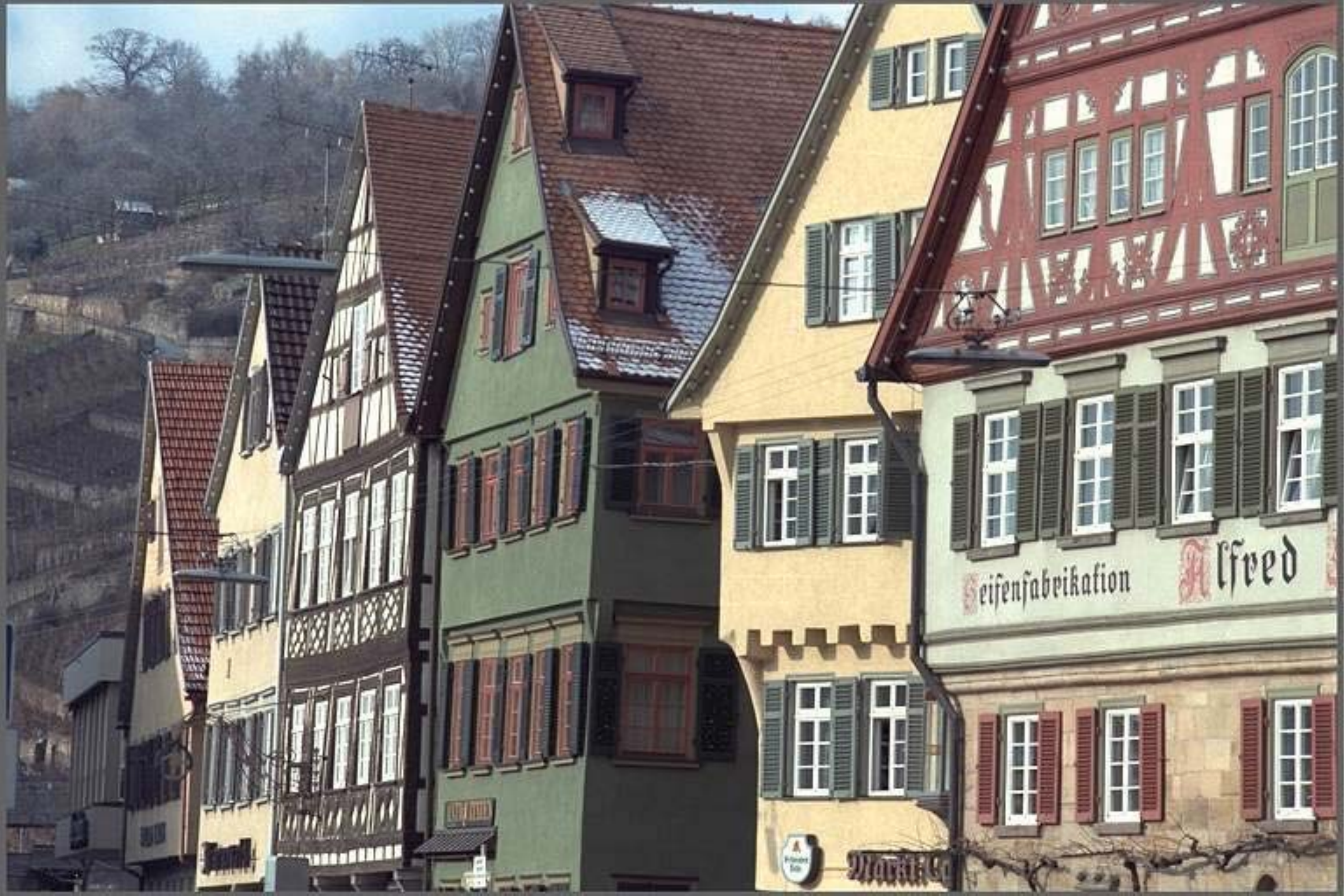}}\\
  \caption{The subjective comparison between two perceptually similar images from LIVE II databases. (a) shows the JPEG version of \textit{house.bmp}, which is compressed at the bit rate of 1.40 bpp. (b) shows the JPEG version of \textit{buildings.bmp}, whose bit rate is 1.78 bpp.}
  \label{fig_img_ref}
\end{figure}

In reviewing these classic indicators \cite{rank_correlation}, there are two rarely mentioned but noteworthy features, i.e., the uniform weighting and hard ranking principles, which assume the sorting mistakes in different quality levels sharing the same significance and each pair of images are sortable. Under these hypotheses, two important perceptual properties are neglected for the IQA task, which easily results a mismatch between the rank correlation evaluation and human preference. Firstly, in many real world applications, the sorting accuracy of high quality images usually plays a more important role than the poor quality ones, where only the top-ranked images are pushed to the users. To illustrate this feature, in Fig. \ref{fig_rank_compare}, we compare two rank results of several denoised images. In this example, there is only one pair of images are mistakenly sorted for both the second and third rows, which are considered to share the same sorting accuracy in terms of the Spearman's Rank Correlation Coefficient (SRCC)  and Kendall's Rank Correlation Coefficient (KRCC). However, as shown in Fig. \ref{fig_rank_compare}, the recommended denoising result from the third row clearly outperforms the one in the second row. When the sorting error occurs in the top-rank, a perceptually poor image is more likely to be pushed to the user as shown in the second row. By contrast, the bottom-rank error for the poor quality images usually shows negligible impact on the recommending results, where the third row still outputs the perceptually best denoising result. Secondly, due to the inherent uncertainty in making judgement \cite{hsu2005neural,Camerer1992}, the ranking results between two images could not provide reliable measurement for the human perception of image quality, when they share the similar quality levels. For clarity, in Fig. \ref{fig_img_ref}, we compare two compressed images whose DMOSs are 13.72 and 13.70, respectively. Let $f(\cdot)$ and $g(\cdot)$ denote two IQA metrics towards DMOS, and we have the evaluation results of $f[\text{Fig. \ref{fig_img_ref} (a)}]>f[\text{Fig. \ref{fig_img_ref} (b)}]$, $g[\text{Fig. \ref{fig_img_ref} (a)}]<g[\text{Fig. \ref{fig_img_ref} (b)}]$. When we follow the hard ranking principle in SRCC and KRCC, it is easy to conclude that $f(\cdot)$ is better than another one $g(\cdot)$. But, in terms of human perception, it is not certain for us to judge whose prediction is better between $f(\cdot)$ and $g(\cdot)$. Particularly, we implement a subjective investigation across 30 human subjects, who are compulsorily required to give a binary response, i.e., Fig. \ref{fig_img_ref} (a) is better or worse than Fig. \ref{fig_img_ref} (b) in terms of image quality. Unsurprisingly, the divisions are serious across all subjects, where 16 votes prefer Fig. \ref{fig_img_ref} (a) and the others disagree with them. The only agreement between all subjects is that it is too hard to make a certain judgement in distinguishing Figs. \ref{fig_img_ref} (a) and (b), which means the hard ranks between perceptually similar images do not contribute to fairly compare different IQA metrics.

Although the two problems mentioned above are widespread in the field of IQA, there are very little researches discussing an alternative rank correlation indicator other than SRCC and KRCC. In \cite{MAD,GMAD}, Wang \textit{et al.} proposed an ingenious maximum differentiation (MAD) solution for comparing different IQA metrics, where each pair of IQA metrics fight with each other by detecting the most noticeable outliers from its competitors. In the following, two new indicators including the aggressiveness and resistance matrixes are designed for the group MAD (gMAD) \cite{GMAD}, which compute the average subjective scores collected from pair comparison. It is worth nothing that the gMAD is initially designed to accelerate the comparison of different IQA algorithms across a large-scale test set. The perceptual importance variance of different quality levels and the subjective uncertainty are not fully considered by the aggressiveness and resistance indicators as well. In addition, the gMAD methodology requires additional manual work for rating the difference of pairwise images in evaluating a new IQA metric, which fairly limits the usability of the aggressiveness and resistance indicators.

To address the aforementioned challenges for IQA metric evaluation, we propose an easy-to-use perceptually weighted rank correlation (PWRC) indicator in this paper. In comparison with the popular SRCC and KRCC, the proposed indicator is characterized by three new features which are summarized in the following:
\begin{enumerate}
  \item \textit{Nonuniform weighting}: To capture the perceptual importance variance in computing the rank correlation, we assign different weights to the rank mistakes occurring in different quality levels. For a pair of images, the weight is jointly determined by the maximum rank of their MOS/DMOS and the difference between their MOS/DMOS. That is, we focus on penalizing great rank mistakes in the high quality levels.
  \item \textit{Selective activation}: In view of the subjective uncertainty for comparing perceptually similar images, we only count the perceptible sorting mistakes for the pairwise images whose MOS/DMOS difference exceeds a given sensory threshold (\textit{ST}). In this way, the proposed indicator could suppress the impact of insensitive sorting errors in measuring the rank correlation between IQA predictions and human-rated MOS/DMOS.
  \item \textit{Graphical representation}: Unlike the traditional numerical representation that shows rank performance with single value, we utilize a graphical plot to illustrate the sorting accuracy (\textit{SA}) variation with a changeable \textit{ST}, which is named by \textit{SA-ST} curve. The graphical representation could compare different IQA metrics more intuitively. Meanwhile, the \textit{SA-ST} curve is capable to measure the superiority of an IQA metric under different confidence intervals, which is beyond both SRCC and KRCC.
\end{enumerate}

Extensive experiments on five publicly available databases show that the proposed PWRC indicator could better distinguish different IQA metrics even they share the same number of mistakenly ranked image pairs. Furthermore, we also validate the applicability of the proposed indicator in recommending robust IQA metrics for both the degraded and enhanced images.

The remainder of this paper is organized as follows. Section II reviews the limitations of the classic SRCC and KRCC statistics. Then, the proposed PWRC indicator is introduced in Section III. We discuss the experimental results in Section IV. Finally, this paper is concluded in Section V.

\section{Rank Correlation Measurement Based on equal weighting}

Given $n$ objects considered referring to two variables $\textbf{x}=\{x_i\}_{i\leq n}$ and $\textbf{y}=\{y_i\}_{i\leq n}$, Kendall \textit{et al.} \cite{rank_correlation} designed a generalized correlation coefficient $\Gamma$ which is given by
\begin{equation}
\Gamma = \frac{\Sigma_{i,j=1}^{n}a_{ij}b_{ij}}{\sqrt{\Sigma_{i,j=1}^{n}a_{ij}^2\Sigma_{i,j=1}^{n}b_{ij}^2}}
\label{eq_gcc}
\end{equation}
where $a_{ij}$ denotes the $x$-correlation variable between $x_i$ and $x_j$, and $b_{ij}$ denotes the $y$-correlation variable between $y_i$ and $y_j$. Both the $a_{ij}$ and $b_{ij}$ are required to be anti-symmetric and satisfy $a_{ij}=-a_{ji}$, $b_{ij}=-b_{ji}$.


By properly setting $a_{ij}$ and $b_{ij}$, we can easily deduce the Spearman's $\rho$ and Kendall's $\tau$ from Eq. (\ref{eq_gcc}). Let $a_{ij}$ and $b_{ij}$ represent the rank deviation of a pair of observation values
\begin{equation}
\begin{aligned}
a_{ij} &= p_i-p_j \\
b_{ij} &= q_i-q_j
\end{aligned}
\end{equation}
where $p_i$ denotes the rank of $x_i$ in \textbf{x}, and $q_i$ denotes the rank of $y_i$ in \textbf{y}. Then, we can obtain $\rho$ as
\begin{equation}
\rho = \frac{\Sigma_{i,j=1}^n(p_i-p_j)(q_i-q_j) }{\sqrt{\Sigma_{i,j=1}^n(p_i-p_j)^2}\sqrt{\Sigma_{i,j=1}^n(q_i-q_j)^2}}
\label{eq_rho_old}
\end{equation}

Assuming there are no tied ranks in \textbf{x} and \textbf{y}, the rank variables $\{p_i\}_{i\leq n}$ and $\{q_i\}_{i\leq n}$ would share equally distributed values ranging in $[1,n]$. Under this assumption, it is easy to deform Eq. (\ref{eq_rho_old}) to a more familiar definition for Spearman's $\rho$ according to \cite{rank_correlation}, i.e.,
\begin{equation}
\rho = 1-w\Sigma_{i=1}^nd_i^2
\label{eq_rho_new}
\end{equation}
where $d_i = i-q_i$, $w=6/(n^3-n)$, and $p_i$ denotes the ground-truth ranks represented by a sequential value $p_i = i$.

It is clear that Spearman's $\rho$ assigns the same weight $w$ to each rank mistake term $d_i^2$. It could reflect the subjective uncertainty in some extent, where an insensitive rank mistake results a smaller $d_i^2$. But, the perceptual importance variance is completely ignored by its uniform weight $w$.

Similarly, when we set $a_{ij}$ and $b_{ij}$ to the binary rank discriminator, i.e.,
\begin{equation}
\begin{aligned}
a_{ij}&=\text{sgn}(p_i-p_j) \\
b_{ij}&=\text{sgn}(q_i-q_j)
\end{aligned}
\end{equation}
we can deduce the Kendall's $\tau$ as
\begin{equation}
\tau=w\sum_{i\neq j=1}^n\text{sgn}(p_i-p_j)\text{sgn}(q_i-q_j)
\label{eq_tau}
\end{equation}
where $w=2/(n^2-n)$, and $\text{sgn}(\cdot)$ is a binary signum function defined by
\begin{equation}
\text{sgn}(p_i-p_j) =
\begin{cases}
1, & p_i-p_j>0 \\
-1, & p_i-p_j<0
\end{cases}
\end{equation}

It is seen that the Kendall's $\tau$ works like a outlier detector, which tries to capture the disagreement of the rank-orders between $(p_i, p_j)$ and $(q_i, q_j)$. Because Eq. (\ref{eq_tau}) only counts the signs of pairwise ranks, it is unable to measure the perceptibility of this outlier, which is determined by the severity of sorting error. Meanwhile, the uniform weight $w$ is also applied to each pair of ranked samples in Eq. (\ref{eq_tau}), which does not consider the perceptual importance variance across different rank levels.

\section{Perceptually Weighted Rank Correlation Indicator}

To better interpret the visual perception behaviors in measuring rank correlation, we propose to assign different weights to each pair of ranked samples by considering both the sorting error degree and the level of the mistaken ranks. Meanwhile, the perceptibility of the mistaken ranks are also identified via a mutable sensory threshold.

Suppose $\textbf{x}=\{x\}_{i\leq n}$ and $\textbf{y}=\{y\}_{i\leq n}$ are the continuous-valued MOSs/DMOSs and IQA predictions, respectively. We first transform them to the discrete rank-orders $\textbf{p}=\{p\}_{i\leq n}$ and $\textbf{q}=\{q\}_{i\leq n}$. Then, the proposed indicator separates the task of rank correlation measurement into three steps including comparison activation, outlier detection, and importance measurement.

Before comparing the agreement between each pair of ($x_i$, $x_j$) and ($y_i$, $y_j$), we need to judge wether this comparison is perceptually sensitive or not. When the MOSs/DMOSs difference between $x_i$ and $x_j$ exceeds the sensory threshold, we will activate this comparison between ($x_i$, $x_j$) and ($y_i$, $y_j$). As discussed in \cite{Swets168,owsley1983contrast,chaieb2008gender,Crawford69,Alais20101362}, the visual sensitivity covers a wide dynamic range across different ages, genders, consciousness, and so on. So, we sample multiple sensory thresholds $\textbf{t}=\{t\}_{k\leq K}$ to evaluate the rank correlation, where the activation function is denoted by $A(\textbf{x}, T)$ and $T=t_k$. In the following, the outlier detection focus on identifying discordant rank-orders between ($p_i$, $p_j$) and ($q_i$, $q_j$), and we denoted it by $D(\textbf{p},  \textbf{q})$. Given all of the activated mistaken rank-orders, we need to assign different weights to them by measuring their perceptual importance. Both the rank-order deviation of (\textbf{p}, \textbf{q}) and the location of the mistaken rank are considered in the importance measurement, which is denoted by $M(\textbf{p}, \textbf{q})$. Finally, by combining these three components together, we can yield an overall sorting accuracy indicator, i.e.,
\begin{equation}
S(\textbf{x}, \textbf{y}, T) = f[A(\textbf{x}, T), D(\textbf{p},  \textbf{q}), M(\textbf{p}, \textbf{q})]
\end{equation}
where $f(\cdot)$ is a combination function for fusing three rank correlation related components.

In defining the rank correlation indicator $S(\textbf{x}, \textbf{y}, T)$, we would like it to possess both the usability and visual perception features, which should satisfy four attributes.

\begin{enumerate}
  \item Symmetry: $S(\textbf{x}, \textbf{y}, T)=S(\textbf{y}, \textbf{x}, T)$;
  \item Boundness: $S(\textbf{x}, \textbf{y}, T)\leq 1$;
  \item Ambiguity: Given a pair of inputs \textbf{x} and \textbf{y}, their rank correlation is non-unique and varies with a changeable sensory threshold $T$;
  \item Unique maximum: $S(\textbf{x}, \textbf{y}, T) = 1$, if and only if \textbf{x} = \textbf{y} and all elements of \textbf{x} satisfy $\|x_i-x_j\|\gg T$ when $i\neq j$.
\end{enumerate}

At first, we define the activation function as
\begin{equation}
A(\textbf{x}, T) =\frac{1}{1+e^{-C_1(\|x_i-x_j\|-T)}}
\label{eq_activation}
\end{equation}
where the constant $C_1$ is used to control the steepness of the activation curve. When $C_1$ grows larger, the activation function would get close to the unit step function.
In comparison with the hard activation by unit step function, the proposed soft activation in Eq. (\ref{eq_activation}) could better model the progressive change of visual sensitivity \cite{visual_field_change,blur_adaption}.

\begin{figure}[t]
  \centering
  \includegraphics[width=0.92\linewidth]{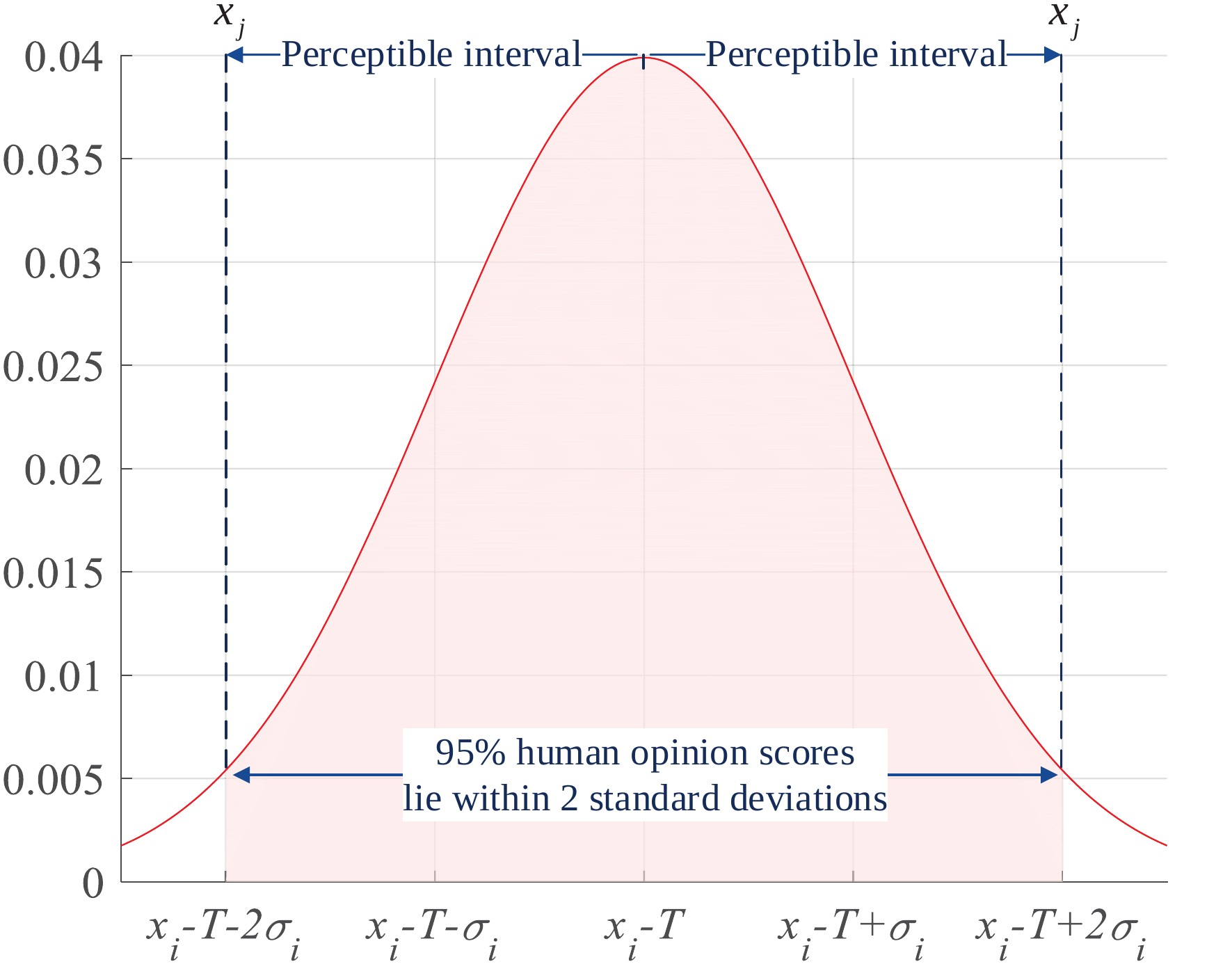}\\
  \vspace{0.14in}
  \caption{The distribution of human opinion scores.}
  \label{fig_GP}
\end{figure}

Following the Gaussian distribution assumption about human opinion score \cite{psychometric}, we consider the MOSs/DMOSs variables satisfy $\textbf{x}\sim\mathcal{N}(x_i,\sigma_i)$ and deduce an appropriate $C_1$ via interval estimation, where $x_i$ and $\sigma_i$ are the mean value and standard deviation of human opinion scores for the $i$th image. Based on the three-sigma rule \cite{three_sigma}, we know that there are about 95\% human opinion scores lying within two standard deviations away from a MOS/DMOS in evaluating each image. In Fig. \ref{fig_GP}, we show the distribution of human opinion scores, which is modeled by a Gaussian distribution. The sensory threshold $T$ adjusts the mean value of this Gaussian distribution to $x_i-T$. When another MOS/DMOS $x_j$ goes beyond $x_i-T$ by the interval of $2\sigma_i$, we believe that these two images are perceived differently at the confidence level of 0.95, which results the equation
\begin{equation}
\frac{1}{1+e^{-C_1\cdot2\sigma_i}}=0.95
\end{equation}
where $C_1$ is approximately equal to $3/(2\sigma_i)$. It is noted that different IQA databases usually collect the MOS/DMOS $x_i$ and standard deviation $\sigma_i$ at different scales. To obtain a general setting for $C_1$, we normalize all subjective scores to the range of [0, 100]. Let $\omega=1/(\max{\{\textbf{x}\}}-\min{\{\textbf{x}\}})$ denote the scaling factor and $\varepsilon=-\min{\{\textbf{x}\}}/(\max{\{\textbf{x}\}}-\min{\{\textbf{x}\}})$ denote the bias term for \textbf{x}.
Then, the normalization for \textbf{x} is defined by
\begin{equation}
\hat{\textbf{x}} =
\begin{cases}
(\omega\textbf{x}+\varepsilon)\times100, & \text{\textbf{x} is MOS}\\
[1-(\omega\textbf{x}+\varepsilon)]\times100, & \text{\textbf{x} is DMOS}
\end{cases}
\label{eq_norm_MOS}
\end{equation}
Meanwhile, all standard deviations $\bm{\sigma}=\{\sigma\}_{i\leq n}$ are normalized by
\begin{equation}
\hat{\bm{\sigma}} =\omega\times \bm{\sigma}\times100
\label{eq_norm_std}
\end{equation}

In this paper, the mean value of $\hat{\bm{\sigma}}$ across three popular IQA databases, i.e., LIVE II \cite{LIVE}, TID2013 \cite{TID2013} and ChallengeDB \cite{ChallengeDB}, is utilized to compute $C_1$, where $E(\hat{\bm{\sigma}})=8.577$ and $C_1=3/(2\times8.577)=0.175$.

The outlier detection function borrows the idea of sign comparison from Kendall's $\tau$
\begin{equation}
D(\textbf{p},  \textbf{q}) =\text{sgn}(p_i-p_j)\text{sgn}(q_i-q_j)
\label{eq_detection}
\end{equation}
where each pair of mistaken rank-orders is labeled by -1. Otherwise, the outlier detection function would denote them by 1. Clearly, more discordant pairs between ($p_i$, $p_j$) and ($q_i$, $q_j$) would lead more negative outputs from Eq. (\ref{eq_detection}).

\begin{figure}[t]
  \centering
  \includegraphics[width=\linewidth]{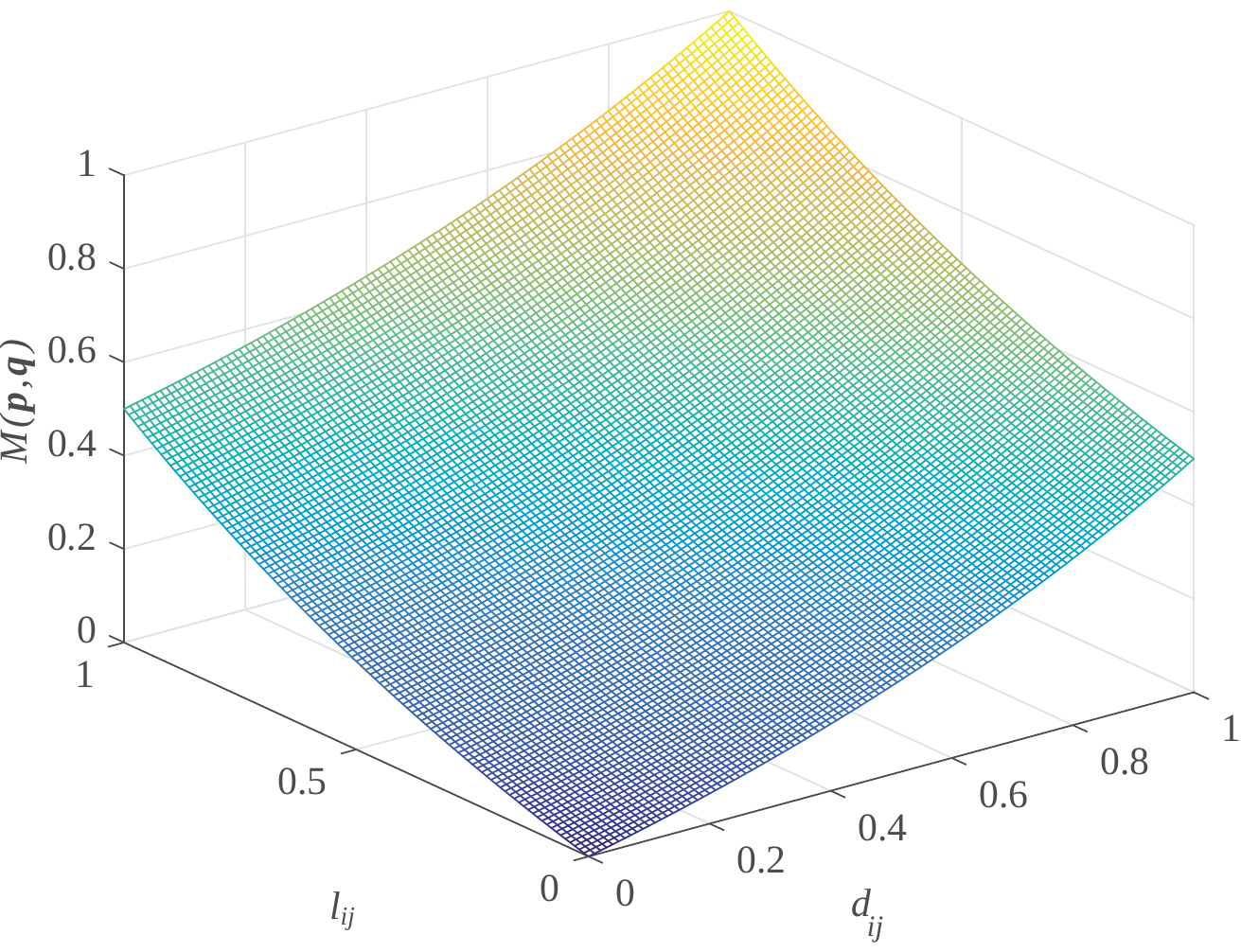}\\
  \caption{The illustration of importance measurement function $M(\textbf{p}, \textbf{q})$.}
  \label{fig_importance}
\end{figure}

In comparing each pair of rank-orders ($p_i$, $p_j$) and ($q_i$, $q_j$), we assign different weights to them by measuring their perceptual importance. Two factors, i.e., rank deviation and rank level, are considered for computing the weight. Particularly, the ground-truth label $p_i$ has been sequentially ranked as $\{p_i = i\}_{i\leq n}$. We represent the normalized rank deviation term between $(p_i, p_j)$ and $(q_i, q_j)$ by $d_{ij}=(\|i-q_i\|+\|j-q_j\|)/(2n-2)$. The normalized rank level term is $l_{ij}=(\max\{i,j\}-1)/(n-1)$. Both $d_{ij}$ and $l_{ij}$ locate in the range [0, 1]. Then, the importance measurement function is defined as
\begin{equation}
M(\textbf{p}, \textbf{q}) = \frac{e^{d_{ij}}+e^{l_{ij}}-2}{\sum_{i\neq j=1}^n (e^{d_{ij}}+e^{l_{ij}}-2)}
\label{eq_importance}
\end{equation}
where $M(\textbf{p}, \textbf{q})\in[0, 1]$ when $n>1$, and $\sum_{i\neq j=1}^n M(\textbf{p}, \textbf{q}) = 1$. Particularly, a larger weight would be assigned to the predicted rank-order $q_i$ which deviates more from a high level ground-truth rank $p_i$. For clarity, the weight variation across different rank deviations and rank levels is shown in Fig. \ref{fig_importance}.

Finally, we combine all three components in Eqs. (\ref{eq_activation}), (\ref{eq_detection}) and (\ref{eq_importance}) to compute the proposed PWRC indicator, i.e.,
\begin{equation}
S(\textbf{x}, \textbf{y}, T) = \sum_{i\neq j=1}^n A(\textbf{x}, T)\cdot D(\textbf{p},  \textbf{q})\cdot M(\textbf{p}, \textbf{q})
\label{eq_PWRC}
\end{equation}
It is clear that the Eq. (\ref{eq_PWRC}) satisfy all four attributes for designing easy-to-use and perception aware rank correlation indicator. Particularly, the Kendall's $\tau$ corresponds to a special case of Eq. (\ref{eq_PWRC}) when we set the comparison activation and importance measurement functions to constants, i.e., $A(\textbf{x}, T)=1$ and $M(\textbf{p}, \textbf{q})=2/(n^2-n)$.

To highlight the difference between PWRC and existing rank correlation indicators, we investigate the evaluation accuracy of different indicators on a set of synthetic data. Assuming that there are five MOS values, i.e., $n=5$, $\textbf{x}=\{x_1=5,x_2=10,x_3=20,x_4=35, x_5=55\}$, and the rank of each MOS is given by $\textbf{p}=\{p_1=1, p_2=2, p_3=3, p_4=4, p_5=5\}$. For comparison with the single-valued indicators SRCC and KRCC, the activation function in PWRC is first set to a constant, i.e., $A(\textbf{x}, T)=1$.
In this investigation, we set the ground-truth rank of \textbf{x} to a descending order permutation $p_5\sim p_1$, which is widely used in kinds of image retrieval and enhancement applications. Then, ten predicted ranks indexed by $\text{S}1\sim \text{S}10$ are evaluated based on different indicators, where the detail results are shown in Table \ref{table_toy_example}. For analysis, the total number of mistaken rank-orders, i.e., $L=\Sigma_{i\neq j=1}^nw[D(\textbf{p},  \textbf{q})]$, is also calculated for each predicted rank, where $w(\cdot)$ is a window function and defined by
\begin{equation}
w[D(\textbf{p},  \textbf{q})] =
\begin{cases}
1, & D(\textbf{p},  \textbf{q})<0\\
0, & \text{otherwise}
\end{cases}
\end{equation}

\begin{table}[t]
  \centering
  \caption{Comparison of different rank correlation indicators}
  \includegraphics[width = \linewidth]{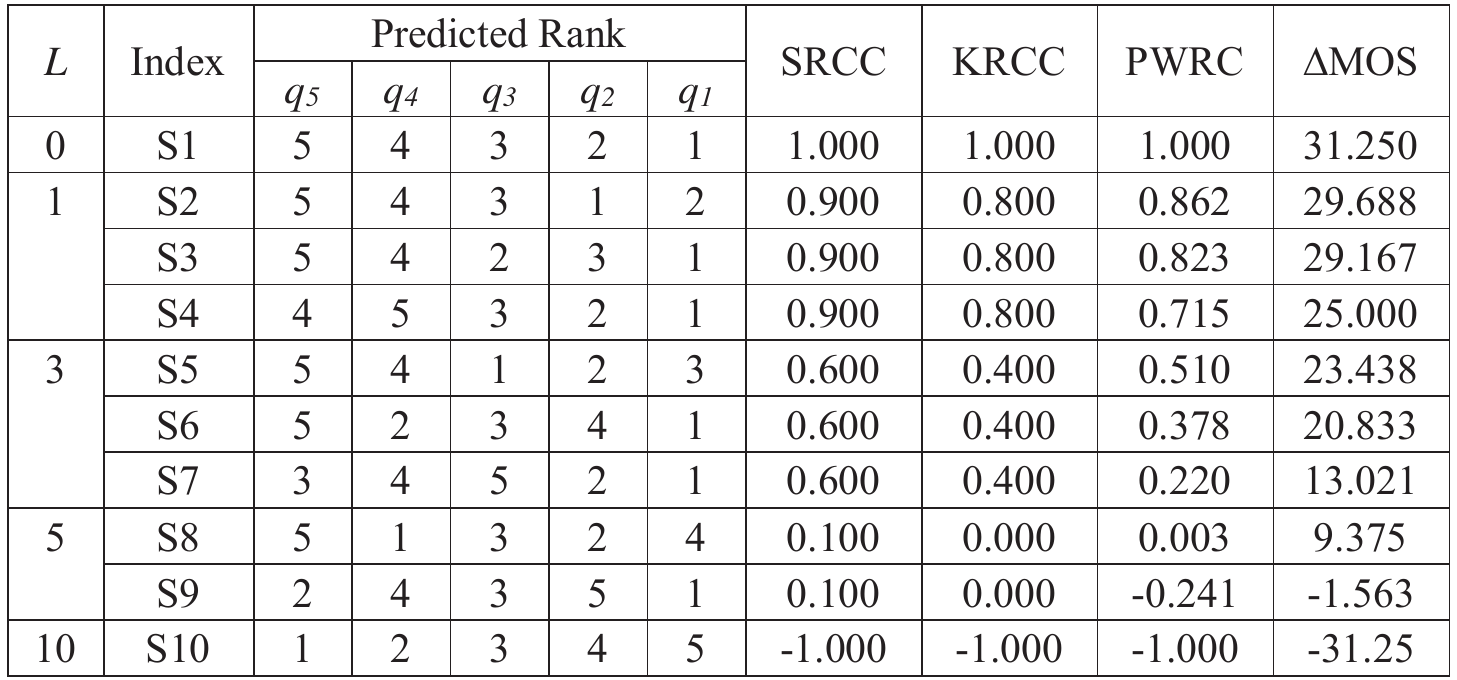}
  \label{table_toy_example}
\end{table}

It is seen that all rank correlation indicators gradually reduce when $L$ increases from 0 to 10. If $L$ is fixed, the SRCC and KRCC would keep the same even the predicted ranks are different. By contrast, the proposed PWRC could further tell the difference between different predicted ranks, which possess the same $L$. To verify the significance of this subtle discriminability, we analyze the average MOS difference between the top \textit{N} ranked samples and the rest ones, which is computed by
\begin{equation}
\triangle d_N = \frac{1}{N}\sum_{i=n+1-N}^{n}x_{q_i}-\frac{1}{n-N}\sum_{j=1}^{n-N}x_{q_j}
\end{equation}
where a larger $\triangle d_N$ means the predicted rank is better for recommending high quality images.
Since $N$ typically varies across different applications, we further calculate the mean value of $\triangle d_N$ by
\begin{equation}
\triangle \text{MOS}= \frac{1}{n-1}\sum_{N=1}^{n-1} \triangle d_N
\label{eq_ave_MOS}
\end{equation}

As shown in Table \ref{table_toy_example}, $\triangle \text{MOS}$ monotonically decreases from the predicted rank S1 to S10. Only the proposed PWRC captures this change, which is crucial for recommending reliable IQA metrics to various quality driven applications.

\begin{figure}[t]
  \centering
  \includegraphics[width=\linewidth]{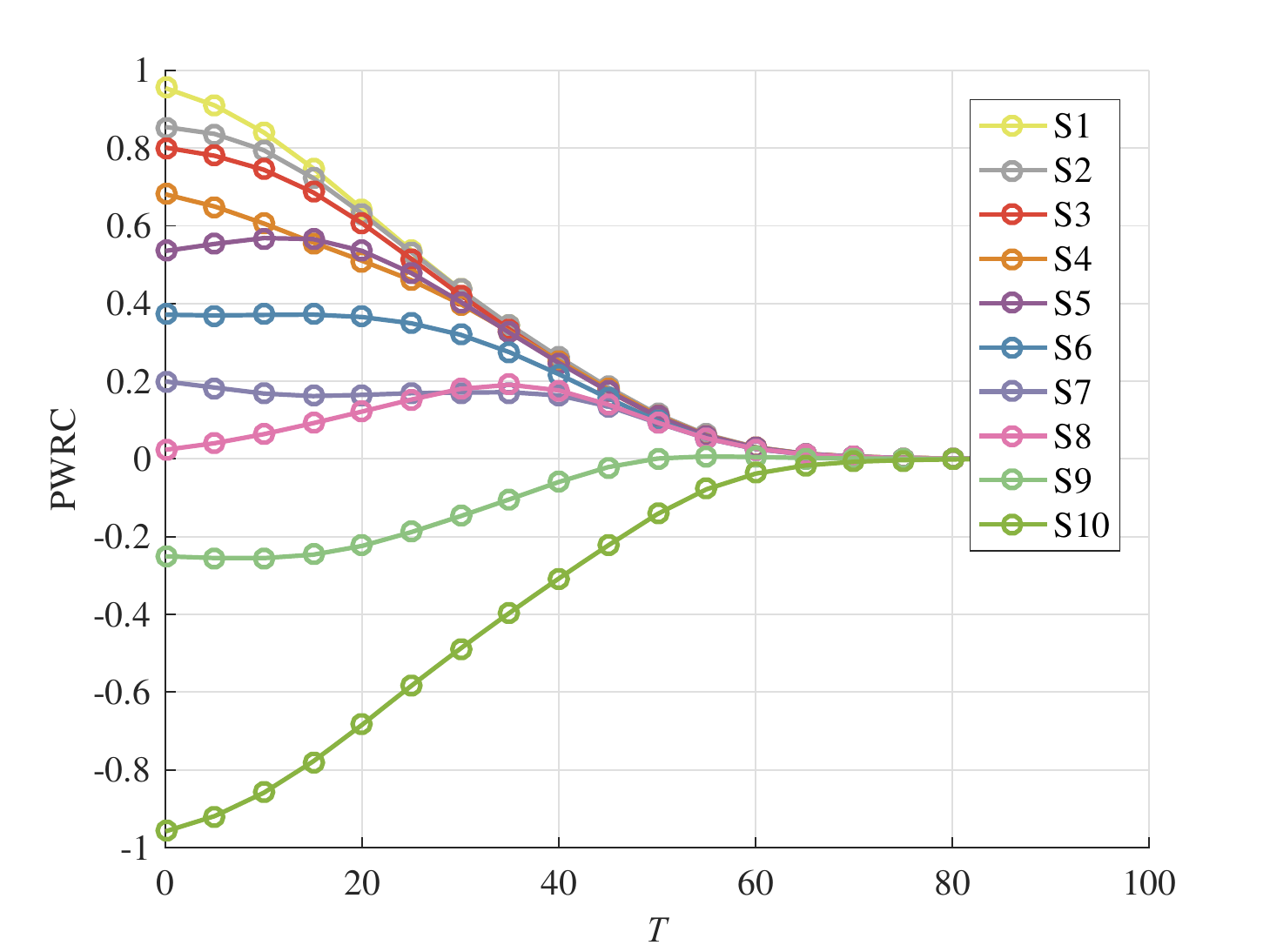}\\
  \caption{The PWRC variation under different sensory thresholds $T$.}
  \label{fig_PWRC_vs_T}
\end{figure}

After the comparison between single-valued indicators, we further illustrate the ambiguity of the proposed PWRC by resorting $A(\textbf{x}, T)$ to Eq. (\ref{eq_activation}). Particularly, twenty evenly spaced sensory thresholds $T$ ranging in [0, 100] are investigated in this section, which produce a \textit{SA-ST} curve for each predicted rank. As shown in Fig. \ref{fig_PWRC_vs_T}, the PWRC indicator dynamically changes when $T$ increases from 0 to 100. Particularly, there are two important features of PWRC revealed from this observation.

Firstly, the PWRC is non-monotonic with respect to an increasing threshold $T$, which may result in different evaluation results in comparing two predicted ranks. For example, in Fig. \ref{fig_PWRC_vs_T}, the predicted rank S4 is superior to S5 when $T$ is smaller than 10. As $T$ increases from 15 to 30, S4 becomes inferior to S5. When $T$ is larger than 35, their performances become close to each other. The similar observations could also be found between S7 and S8. The main reasons behind this observation lie in the contrary effects of the inactivation $A(\textbf{x}, T)$ towards different responses in $D(\textbf{p},  \textbf{q})$ and the nonuniform weighting $M(\textbf{p}, \textbf{q})$ for different quality levels. More specifically, given two pairs of correctly and mistakenly ranked images, their inactivations would bring PWRC reduction and elevation, respectively. In addition, when the inactivation is applied to high quality image pairs, the PWRC reduction or elevation would be more significant. Otherwise, the PWRC change would be relatively small. This feature enables PWRC to evaluate an IQA metric more comprehensively, where a changeable threshold $T$ could capture the visual sensitivities across different genders, ages and so on \cite{Swets168,owsley1983contrast}.

\begin{figure}[t]
  \centering
  \includegraphics[width=\linewidth]{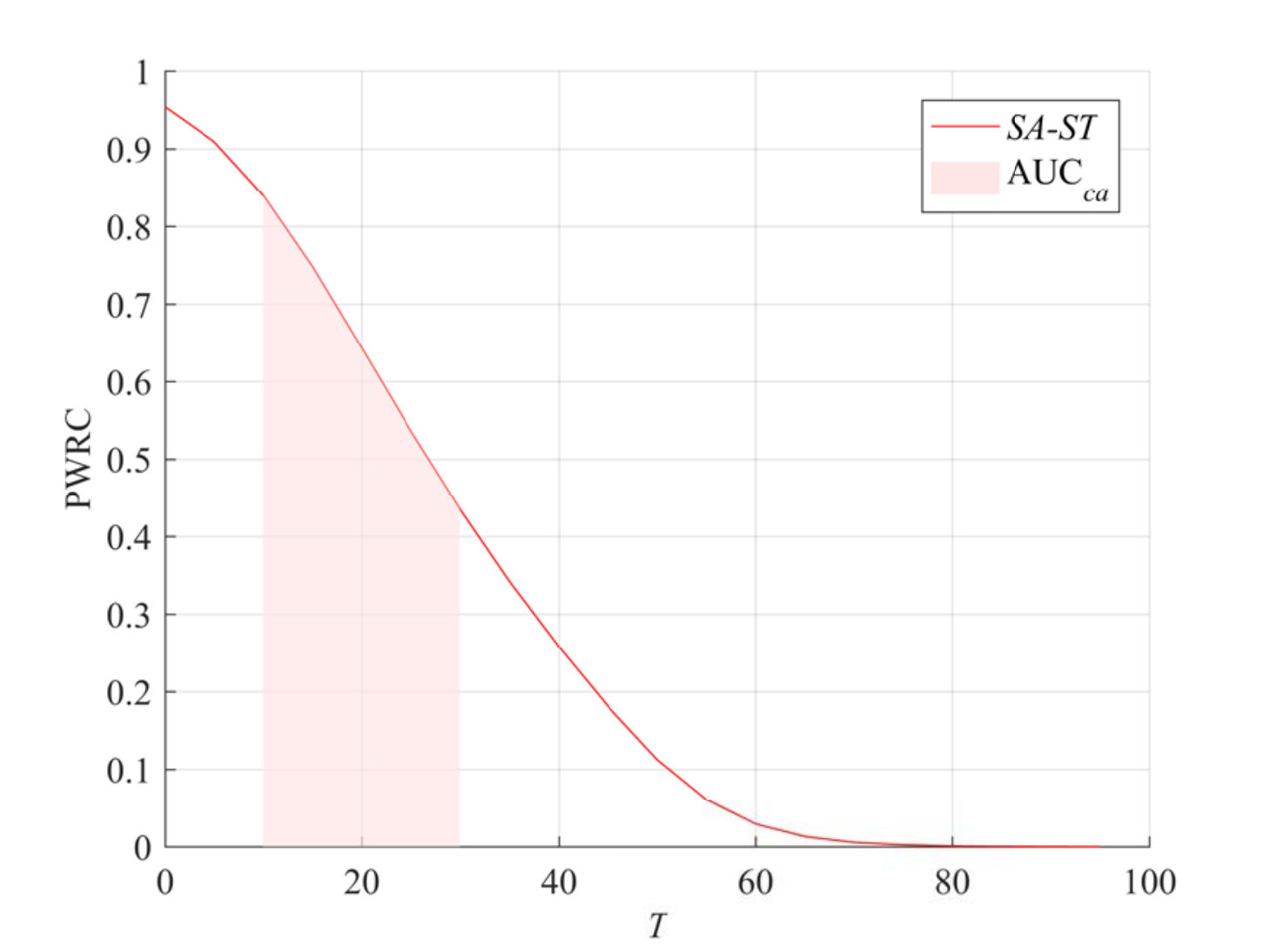}\\
  \caption{The $\text{AUC}_{ca}$ computed under a given sensory threshold range.}
  \label{fig_AUC}
\end{figure}

Secondly, although the PWRC shows different variation tendencies for various predicted ranks, they tend to approach each other with an increasing $T$. This feature enhances the fault-tolerant capability of PWRC with respect to unsortable images, whose MOSs/DMOSs are close to each other. More specifically, if an IQA metric's PWRC outperforms the others on a larger $T$, the confidence of its superiority would be higher. By accumulating PWRC across a given threshold range [$T_{min}$, $T_{max}$], we derive a confidence-aware rank correlation measurement from the area under the curve (AUC), which is defined by
\begin{equation}
\text{AUC}_{ca}=\int_{T_{min}}^{T_{max}}S(\textbf{x}, \textbf{y}, T)dT
\label{eq_AUC}
\end{equation}
where $T_{min}=\min\{2\hat{\bm{\sigma}}\}$ and $T_{max}=\max\{2\hat{\bm{\sigma}}\}$. $\hat{\bm{\sigma}}$ is the set of all normalized standard deviations associated to the MOSs/DMOSs in each IQA database. For clarity, an illustration of $\text{AUC}_{ca}$ is shown in Fig. \ref{fig_AUC}.

\section{Experiments}
\subsection{Protocols}
In this section, we investigate the performance of different rank correlation indicators on five publicly available IQA databases, which include the LIVE II \cite{LIVE}, TID2013 \cite{TID2013}, ChallengeDB \cite{ChallengeDB}, IVCDehazing \cite{IVCDehazing} and ESPL-LIVE \cite{HDR}. Particularly, each database explores the human perception of image quality towards different visual contents. The LIVE II and TID2013 databases collect the human opinion scores on thousands of images contaminated by artificially simulated distortions, such as, JPEG2000, JPEG, additive Gaussian white noise, Gaussian blur, fast fading and so on. In ChallengeDB, there are 1162 authentically distorted images collected from a wide variety of mobile camera devices including the smart phones and tablets. The IVCDehazing database focuses on studying the IQA problem for the enhanced image under foggy weather. For the ESPL-LIVE database, a large-scale subjective scores are collect on 1811 high dynamic range (HDR) images which are created by different tone mapping and multi-exposure fusion techniques.

For comparison, we use different rank correlation indicators to evaluate thirteen popular IQA algorithms, which include six full-reference (FR) metrics, i.e., PSNR \cite{MSE}, IWPSNR \cite{info_weighting}, SSIM \cite{SSIM}, IWSSIM \cite{FR_IQA}, MSSSIM \cite{msssim}, FSIM \cite{FSIM}, and seven no-reference (NR) metrics, i.e., BIQI \cite{BIQI}, BLIINDS II \cite{BLIINDS-II}, BRISQUE \cite{BRISQUE}, DIIVINE \cite{DIIVINE}, NFERM \cite{NFERM}, M3 \cite{BIQA_GM_LOG} and TCLT \cite{TCLT}.

Since the NR-IQA algorithms require additional training process, we follow the random split criteria in \cite{BIQI,BLIINDS-II,DIIVINE,BRISQUE,NFERM,BIQA_GM_LOG,TCLT} and separate each IQA database into non-overlapped training and testing sets over 1000 times. More specifically, in each trial, we randomly choose part of reference images and their contaminated or enhanced versions to construct the training set. The testing set is composed of the rest images, whose visual contents are different from the training set. It is noted that all images collected in ChallengeDB database do not contain any reference images or overlapped visual contents. So, we straightforwardly divide it to two parts for training and testing, respectively. In addition, all FR-IQA algorithms are evaluated on the same test set like the other NR-IQA metrics, which ensures a fair comparison under consistent experimental setup.

In this section, three split ratios are investigated where the training set takes up 80, 50, and 20 percents of images in each database. The median values of SRCC, KRCC and $\text{AUC}_{ca}$ across 1000 random split trials are used to rank different IQA algorithms. Since the quality-driven applications prefer an accurate push capability for high quality images, we employ $\Delta \text{MOS}$ as the benchmark for ranking different IQA metrics. Given a rank correlation indicator, the disagreements between its rank results and $\Delta \text{MOS}$ are used for measuring the push accuracy, where more disagreements mean worse accuracy.

\subsection{Implementation Details}
As discussed in Section III, we compute the proposed PWRC on normalized MOS/DMOS values and standard deviations. Given an IQA database, the scaling factor $\omega$ and bias term $\varepsilon$ are computed on the set of all MOS/DMOS values. Both the ground-truth subjective scores and the predicted image qualities are normalized by Eqs. (\ref{eq_norm_MOS}) and (\ref{eq_norm_std}). Then, we compute the confidence-aware indicator $\text{AUC}_{ca}$ from PWRC curve to rank different IQA metrics, where a sensory threshold range [$T_{min}$, $T_{max}$] is required as shown in Eq. (\ref{eq_AUC}). It is noted that different IQA databases would produce different $\omega$, $\varepsilon$ and [$T_{min}$, $T_{max}$]. For clarity, we summarize the parameters of different databases in Table \ref{table_param_summary}, where a larger MOS/DMOS standard deviation would require wider sensory threshold range for computing $\text{AUC}_{ca}$.

\begin{figure*}[t]
\centering
  \subfigure[LIVE II database with 80\% training data]{
  \includegraphics[width=.31\linewidth]{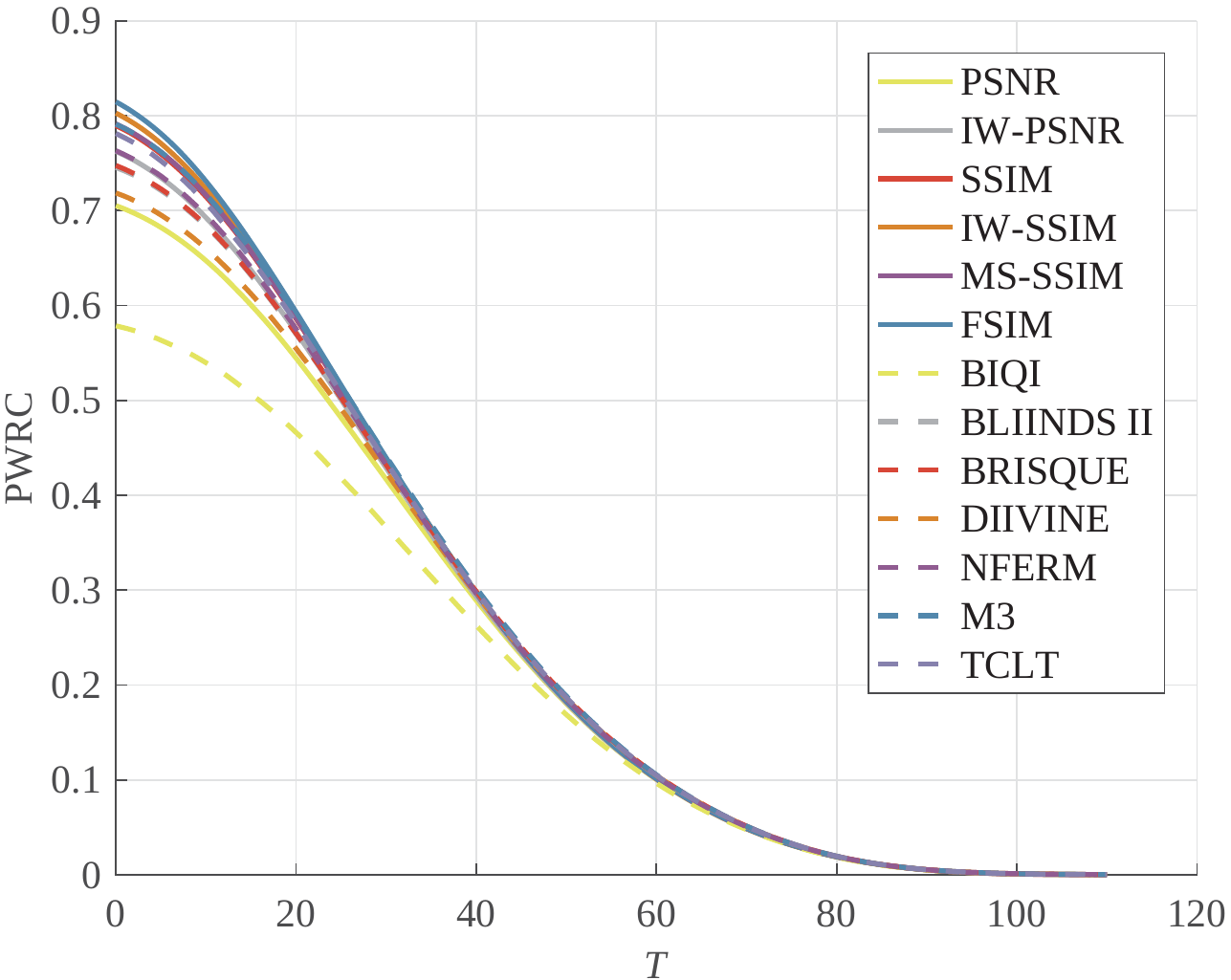}}
  \subfigure[TID2013 database with 80\% training data]{
  \includegraphics[width=.31\linewidth]{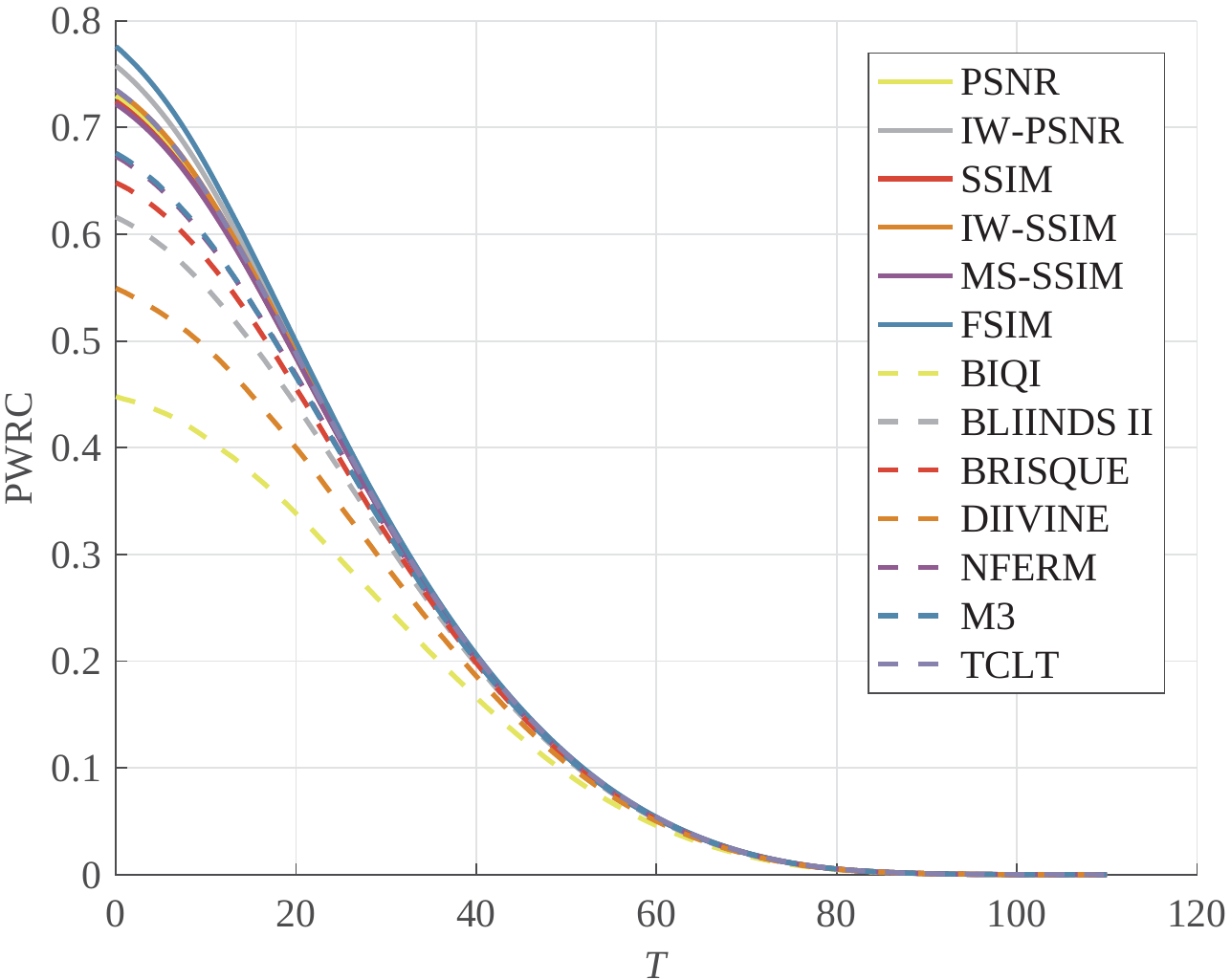}}
  \subfigure[ChallengeDB database with 80\% training data]{
  \includegraphics[width=.31\linewidth]{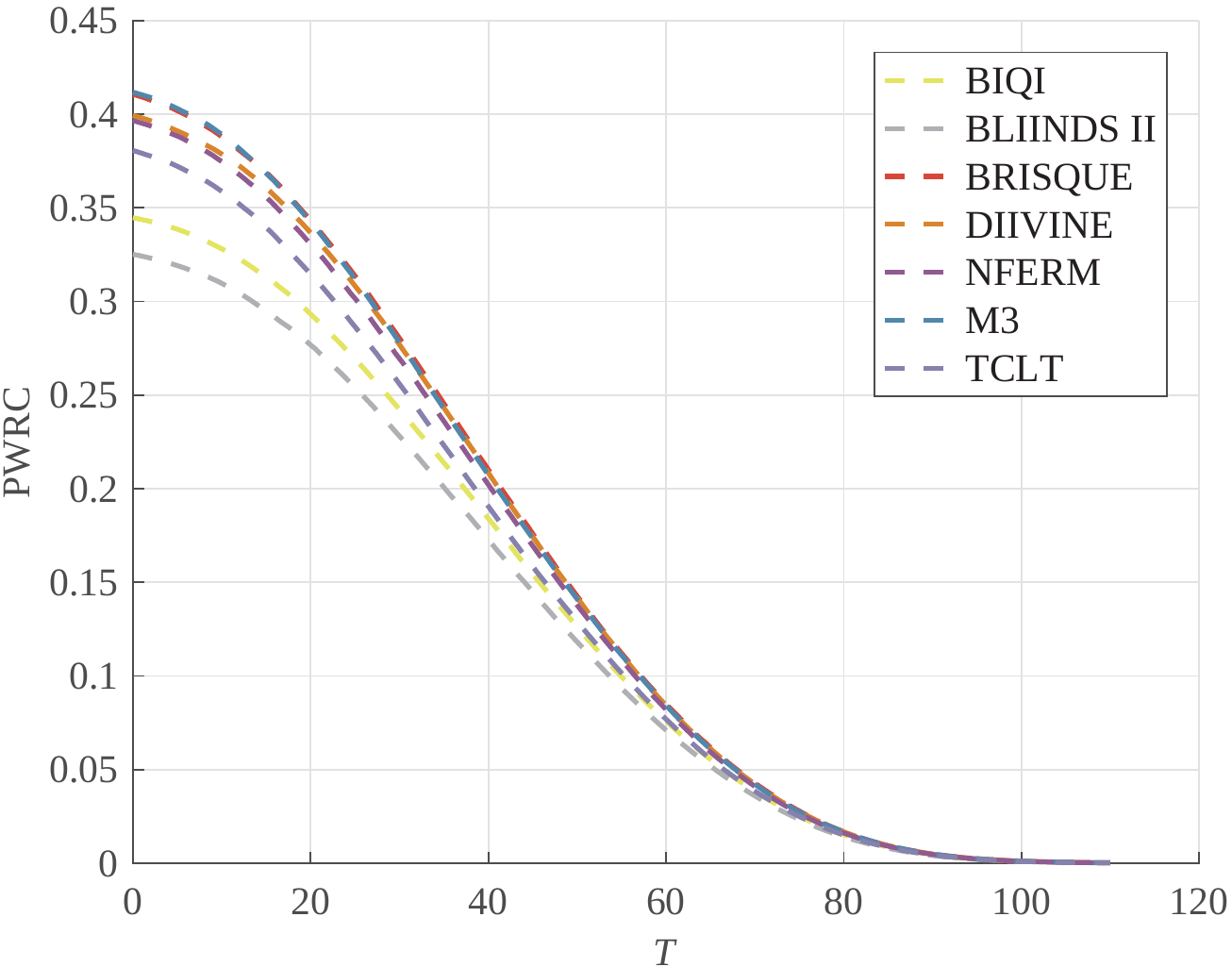}}
  \subfigure[LIVE II database with 50\% training data]{
  \includegraphics[width=.31\linewidth]{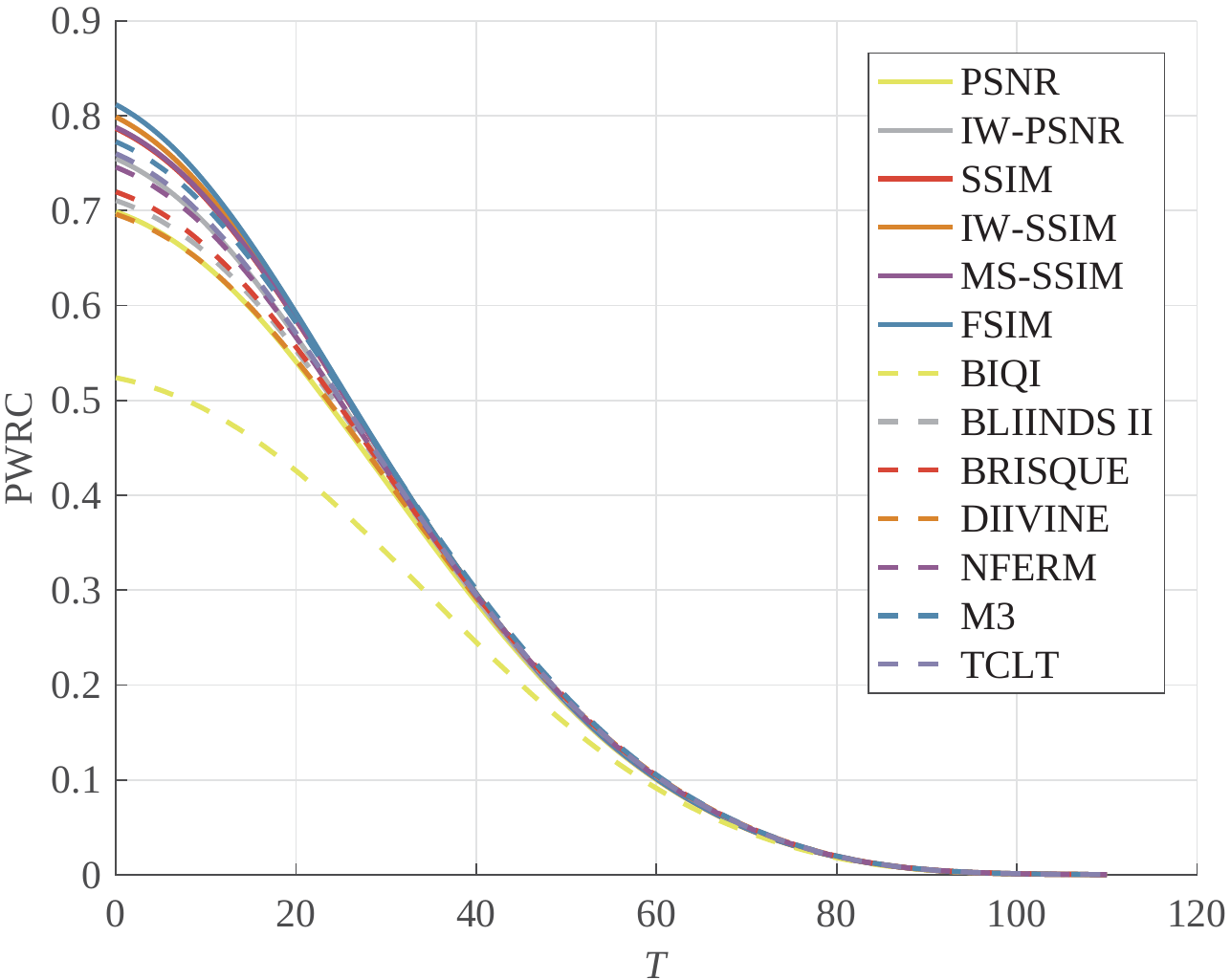}}
  \subfigure[TID2013 database with 50\% training data]{
  \includegraphics[width=.31\linewidth]{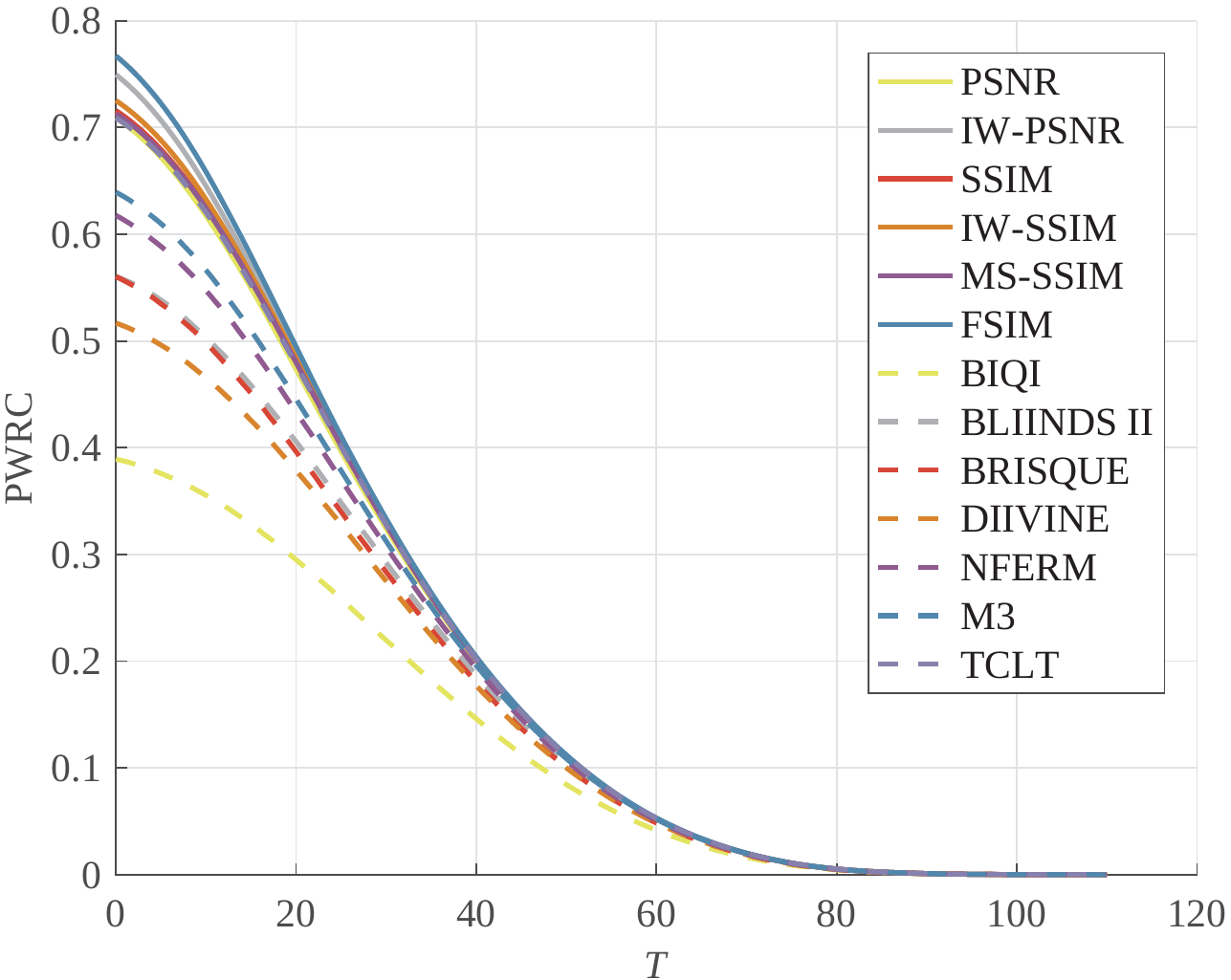}}
  \subfigure[ChallengeDB database with 50\% training data]{
  \includegraphics[width=.31\linewidth]{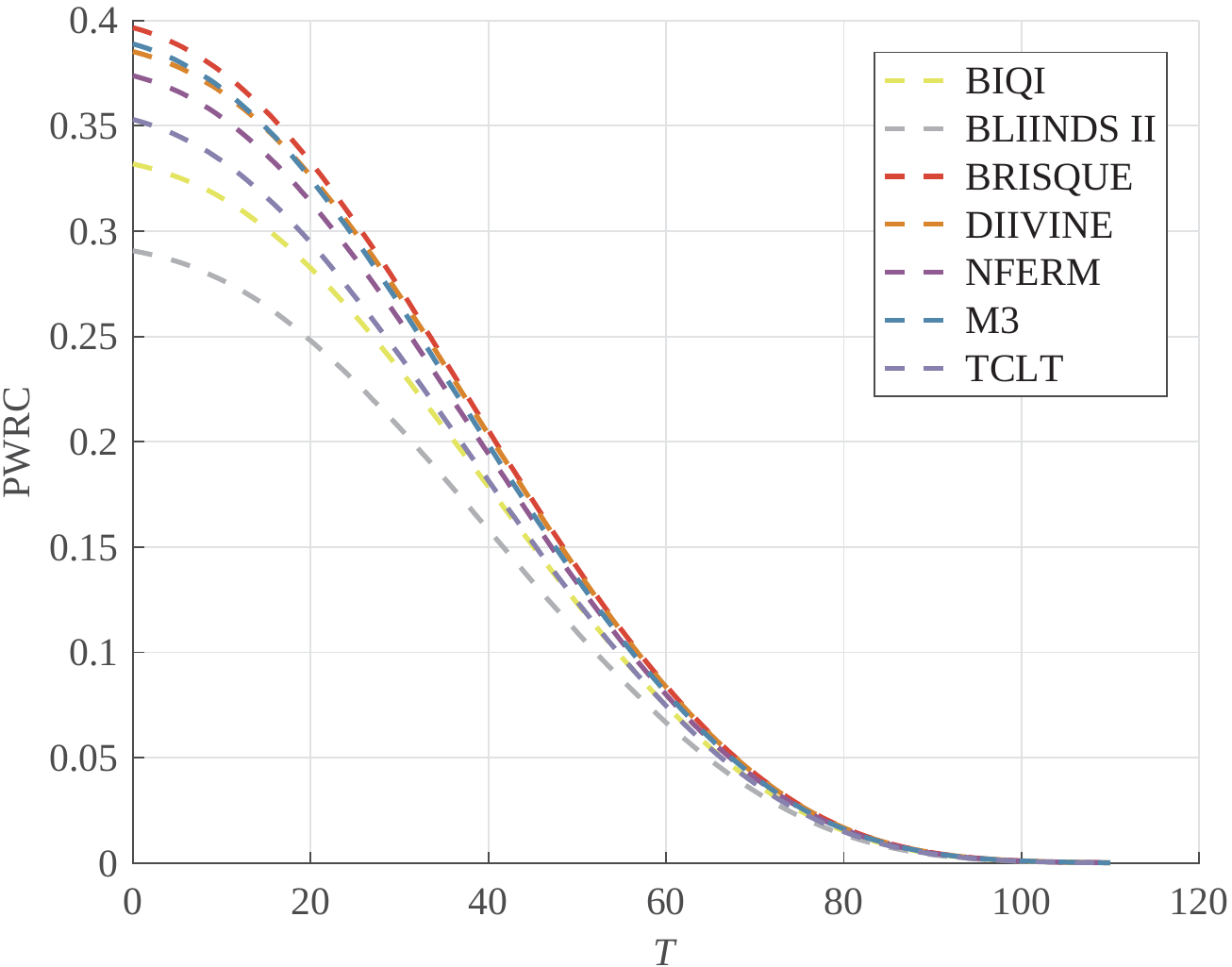}}
  \subfigure[LIVE II database with 20\% training data]{
  \includegraphics[width=.31\linewidth]{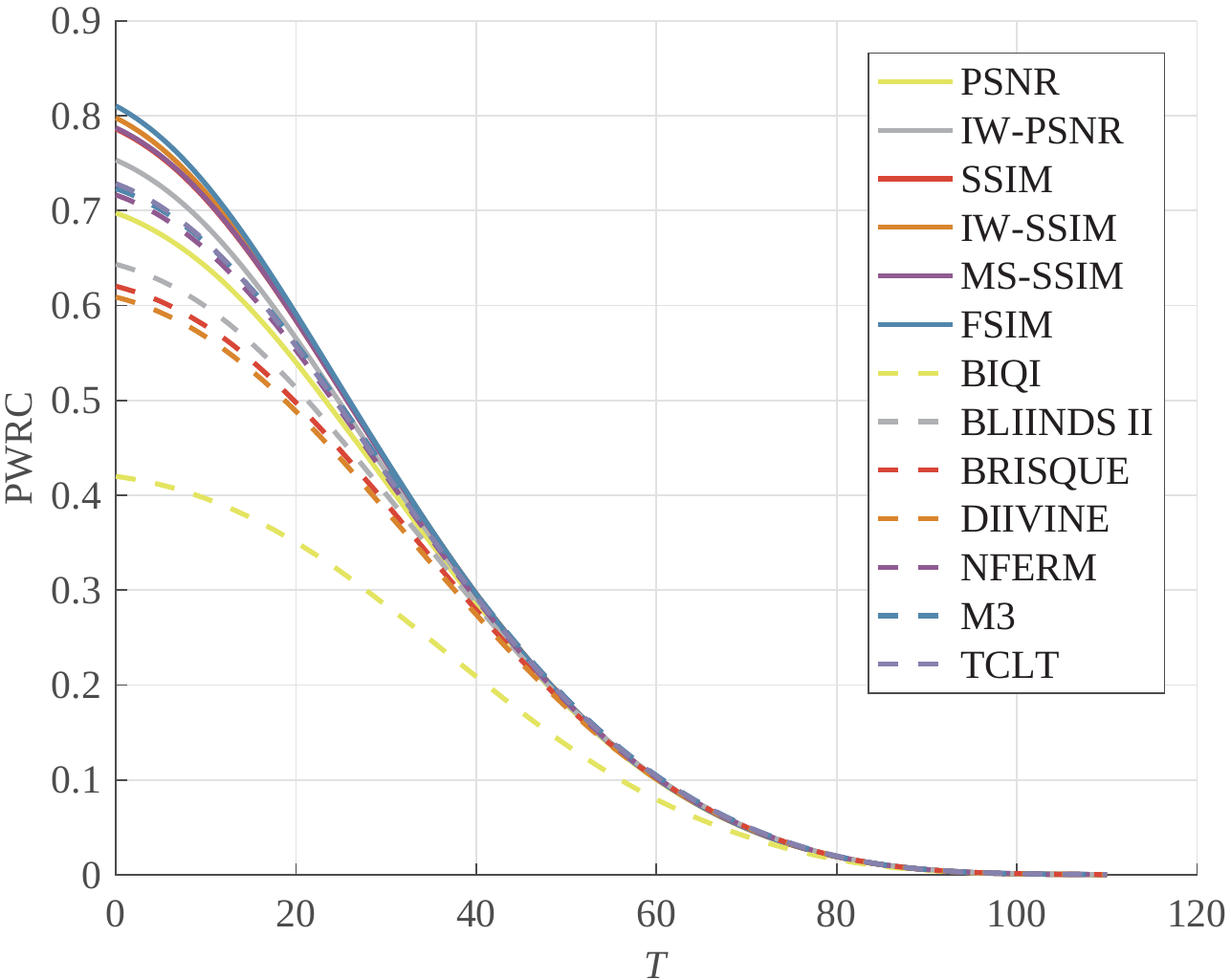}}
  \subfigure[TID2013 database with 20\% training data]{
  \includegraphics[width=.31\linewidth]{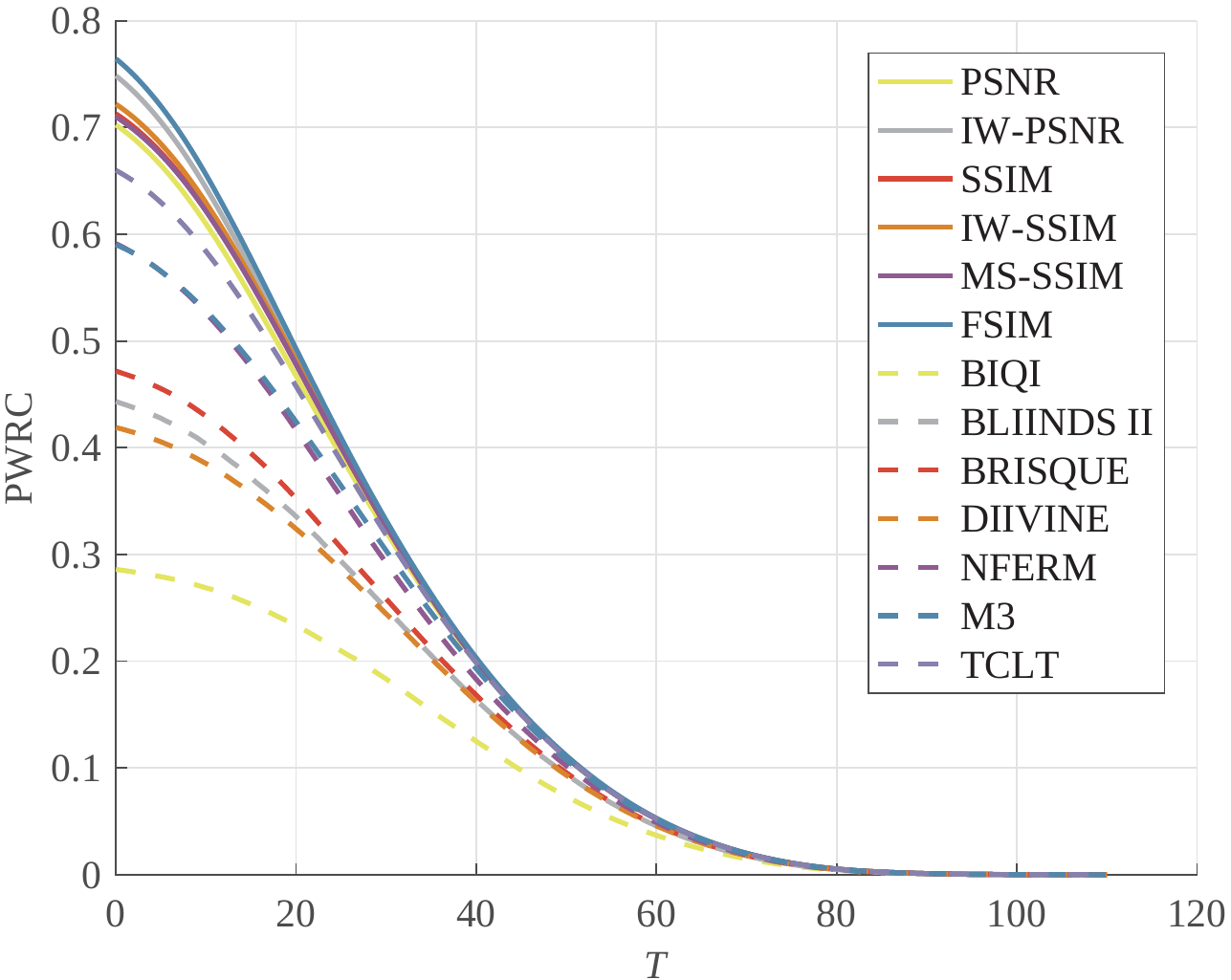}}
  \subfigure[ChallengeDB database with 20\% training data]{
  \includegraphics[width=.31\linewidth]{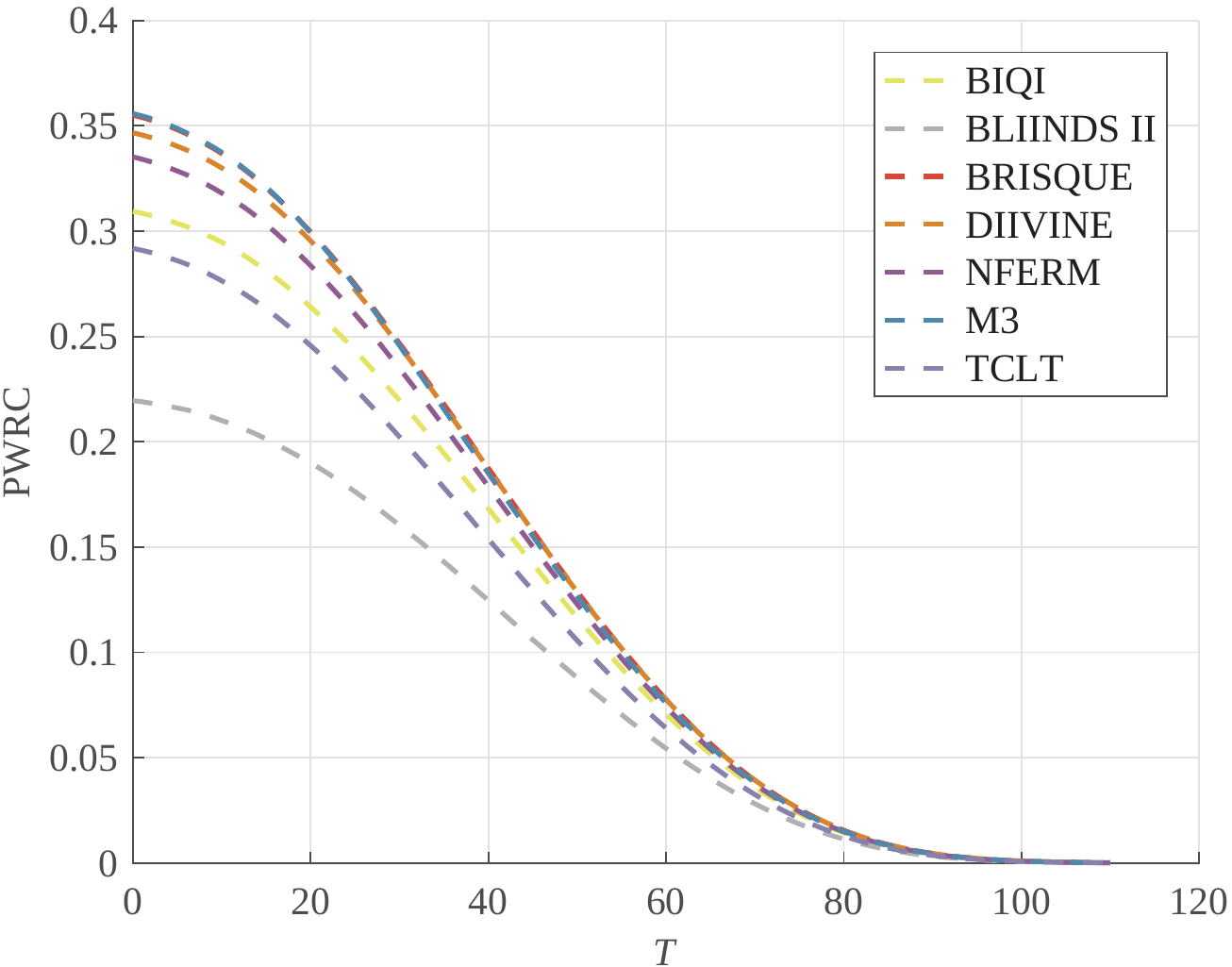}}
  \caption{The \textit{SA-ST} curves of different IQA algorithms tested on three databases. (a)-(c) show the PWRC performances on LIVE II, TID2013, and ChallengeDB, respectively, where the training sets take up 80\% images in each database. (d)-(f) give the rank correlation results whose training sets occupy 50\% images. (g)-(i) report the ranking accuracy when training sets occupy 20\% images.}
  \label{fig_SAST_curves}
\end{figure*}

\begin{table}[t]
  \centering
  \caption{The parameters for PWRC on different databases}
  \includegraphics[width = \linewidth]{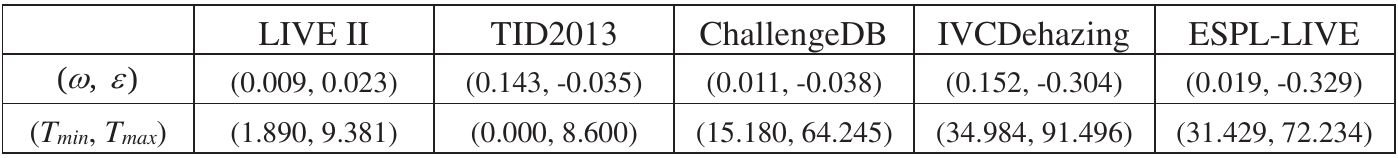}
  \label{table_param_summary}
\end{table}

\subsection{Evaluation on Degraded Images}

\begin{table*}[t]
  \centering
  \caption{Comparison of different rank correlation indicators on LIVE II database}
  \includegraphics[width = .8\linewidth]{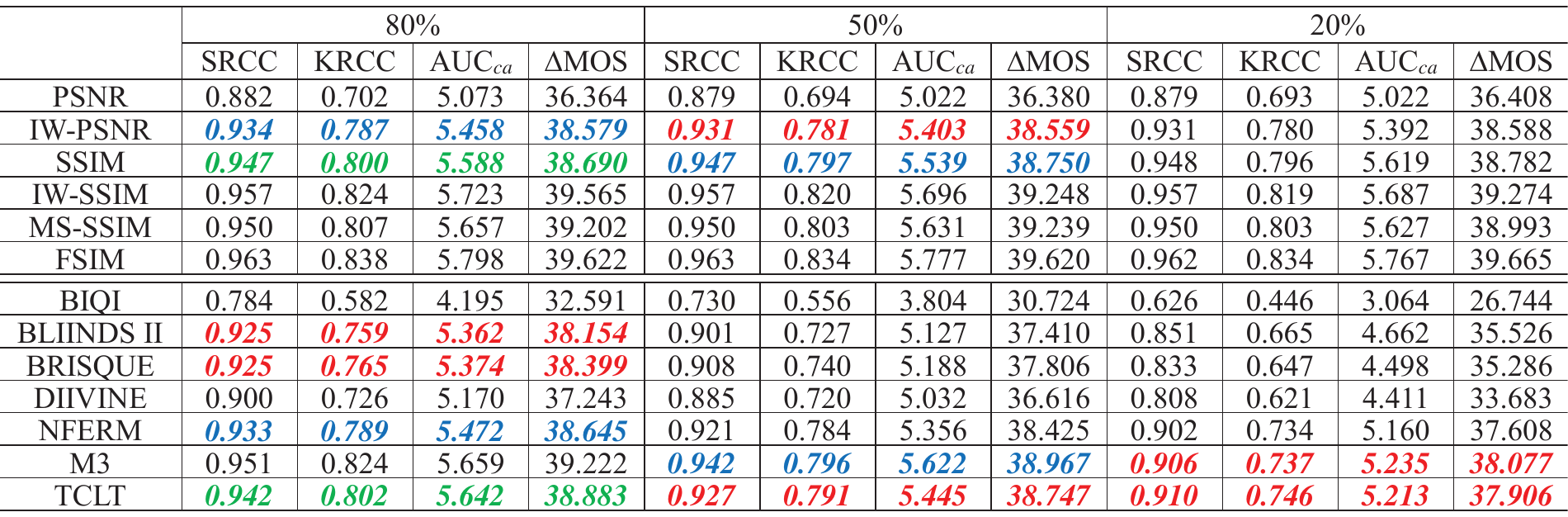}
  \label{table_LIVE_results}
\end{table*}

\begin{table*}[t]
  \centering
  \caption{Comparison of different rank correlation indicators on TID2013 database}
  \includegraphics[width = .8\linewidth]{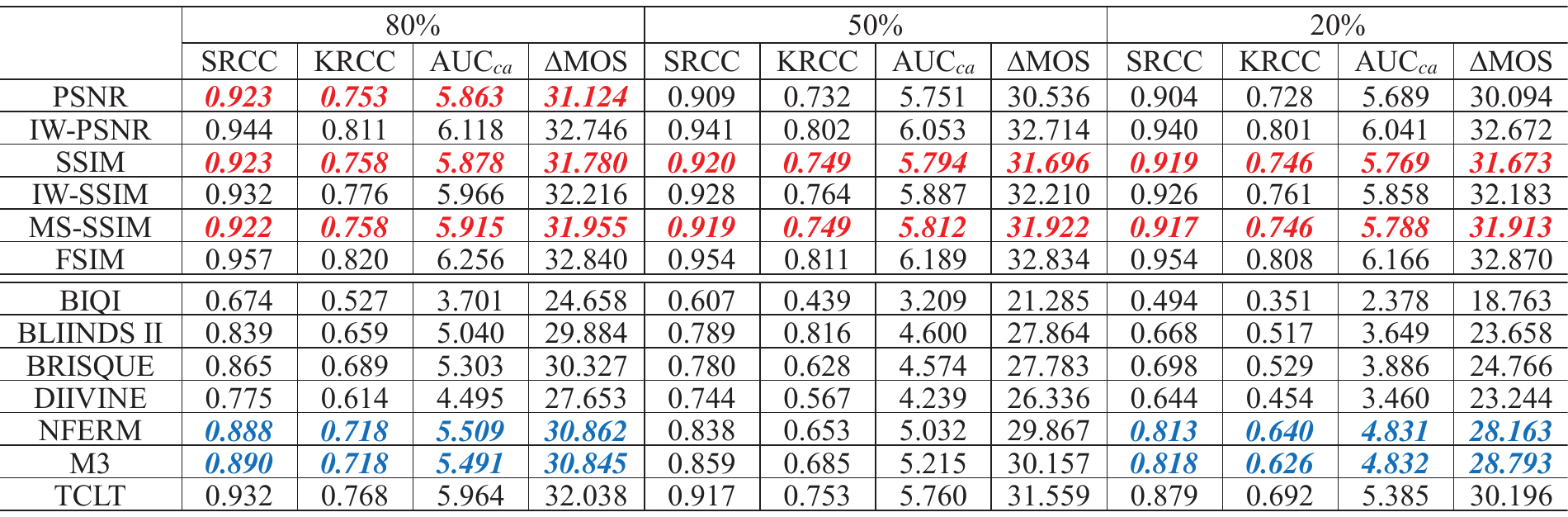}
  \label{table_TID2013_results}
\end{table*}

\begin{table*}[t]
  \centering
  \caption{Comparison of different rank correlation indicators on ChallengeDB database}
  \includegraphics[width = .8\linewidth]{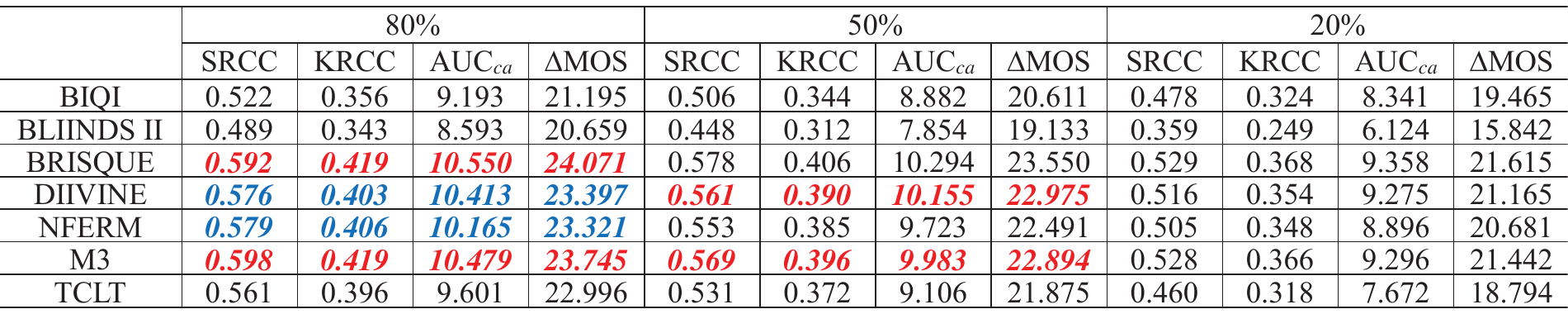}
  \label{table_ChallengeDB_results}
\end{table*}

In this section, we first employ the proposed PWRC to evaluate the rank performance of different IQA metrics towards degraded images, which are investigated on LIVE II, TID2013 and ChallengeDB databases, respectively. To illustrate an intuitive comparison results, the \textit{SA-ST} curves of different IQA algorithms are shown on Fig. \ref{fig_SAST_curves}. It is noted that the ChallengeDB database does not contain uncontaminated reference images. So only the results of NR-IQA algorithms are reported in Fig. \ref{fig_SAST_curves}. It is seen that existing IQA metrics work well on artificially simulated distortions, where most PWRC results are larger than 0.5 on LIVE II and TID2013 databases. However, in coping with the authentically distorted images, the performances of all IQA metrics are very poor, whose PWRC results are all smaller than 0.4 on the ChallengeDB database. In addition, it is found that the performances of the training based NR-IQA algorithms would gradually decrease when the ratio of training set reduces from 80\% to 20\%. By contrast, the FR-IQA metrics achieve more robust results, where the similar results could also be found in \cite{GMAD}. As shown in Fig. \ref{fig_SAST_curves} (a), all NR-IQA algorithms, apart from BIQI, achieve comparable PWRC performance in comparison with the FR-IQA metrics when the training set takes up 80\% images. While, in Fig. \ref{fig_SAST_curves} (g), the FR-IQA algorithms outperform most NR-IQA algorithms when the training set ratio drops to 20\%.

To quantitatively compare the rank results towards different IQA metrics, we further report their SRCC, KRCC, $\text{AUC}_{ca}$ and $\Delta \text{MOS}$ values on LIVE II, TID2013 and ChallengeDB databases, which are shown in Tables \ref{table_LIVE_results}, \ref{table_TID2013_results} and \ref{table_ChallengeDB_results}. Particularly, we highlight all pairwise metrics by boldface and italic when they present inconsistent rank results between $\Delta \text{MOS}$ and the other indicators. Each mistaken pair is labeled by the same color, such as, red, blue and green. As shown in Table \ref{table_LIVE_results}, when 80\% images are used for training on the LIVE II database, the SRCC considers BLIINDS II and BRISQUE present the same rank accuracy, whose reported values are both 0.925. But, the $\Delta \text{MOS}$ value of BRISQUE is clearly better than BLIINDS II. Both the KRCC and the proposed $\text{AUC}_{ca}$ report the consistent rank result with respect to $\Delta \text{MOS}$. When IW-PSNR is compared with NFERM, the SRCC prefers the FR-IQA metric IW-PSNR, which is also opposite to the $\Delta \text{MOS}$ and the other indicators. When the training set ratio is set to 50\%, both the SRCC and KRCC would recommend SSIM in comparison with M3. However, we can find that M3 achieves higher $\Delta \text{MOS}$, and the proposed $\text{AUC}_{ca}$ indicator correctly reflects this superiority. The similar result is also reported in comparing M3 with TCLT when we use 20\% images for training. Meanwhile, more disagreements between $\Delta \text{MOS}$ and SRCC/KRCC could be found in Tables \ref{table_TID2013_results} and \ref{table_ChallengeDB_results}.

Across all of experiments on LIVE II, TID2013 and ChallengeDB databases, we can find that the $\text{AUC}_{ca}$ computed from our proposed PWRC achieves highly consistent rank results with respect to $\Delta \text{MOS}$, where there is no disagreement between them. It verifies that the proposed PWRC indicator is beneficial for pushing perceptually preferred images from the degraded data, where the top ranked images possess higher average MOS.

\subsection{Evaluation on Enhanced Images}

\begin{figure*}[t]
\centering
  \subfigure[IVCDehazing database with 80\% training data]{
  \includegraphics[width=.31\linewidth]{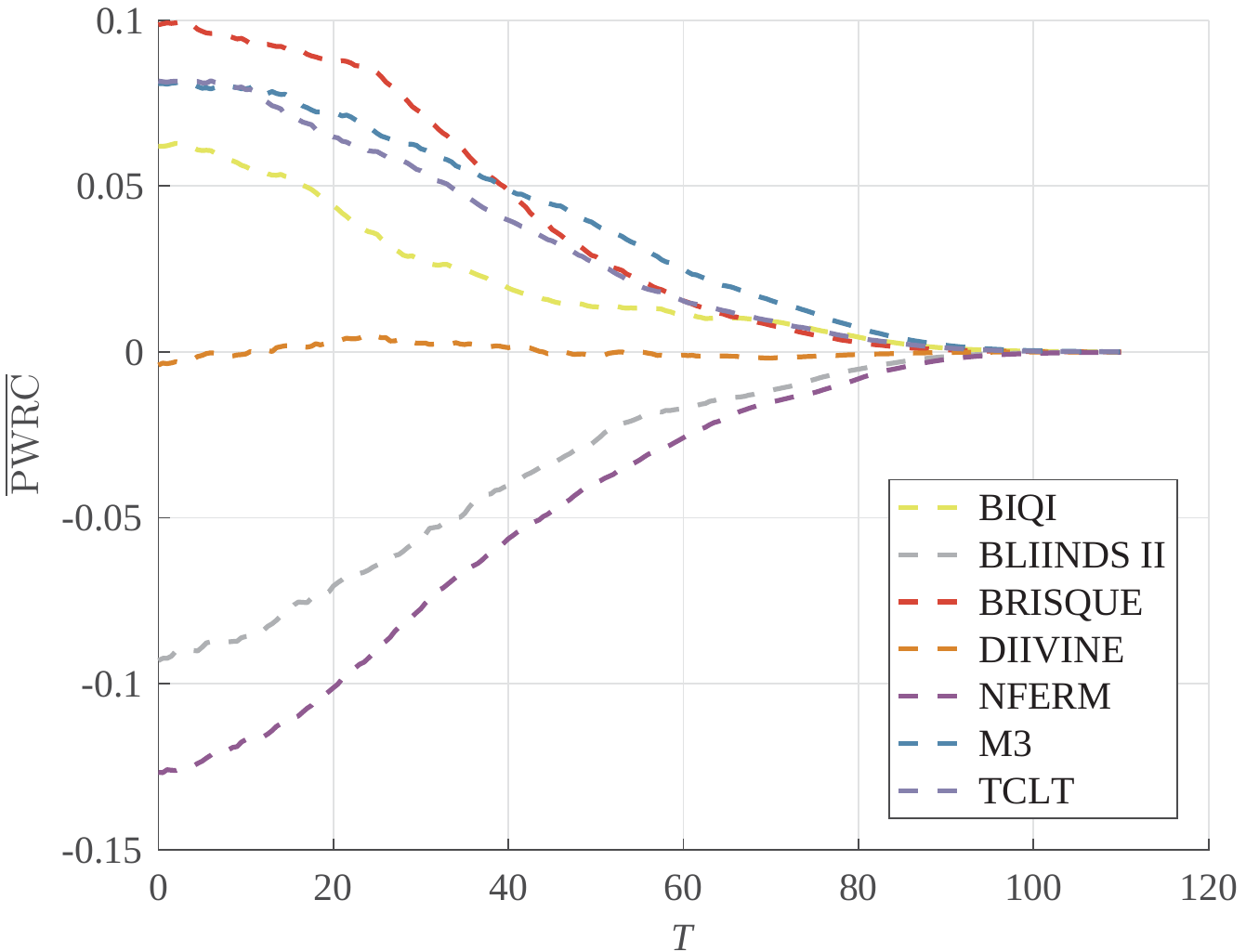}}
  \subfigure[IVCDehazing database with 50\% training data]{
  \includegraphics[width=.31\linewidth]{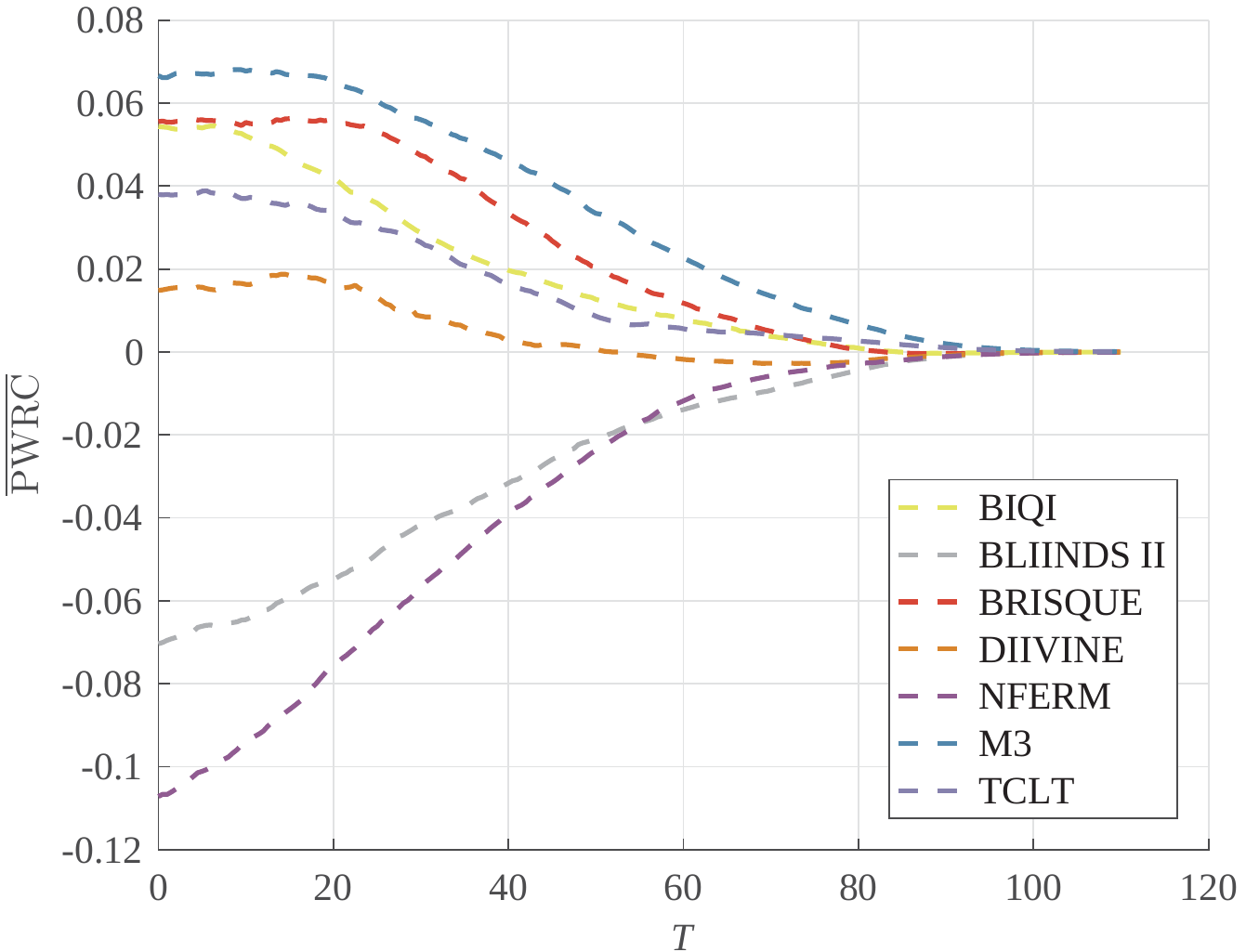}}
  \subfigure[IVCDehazing database with 20\% training data]{
  \includegraphics[width=.31\linewidth]{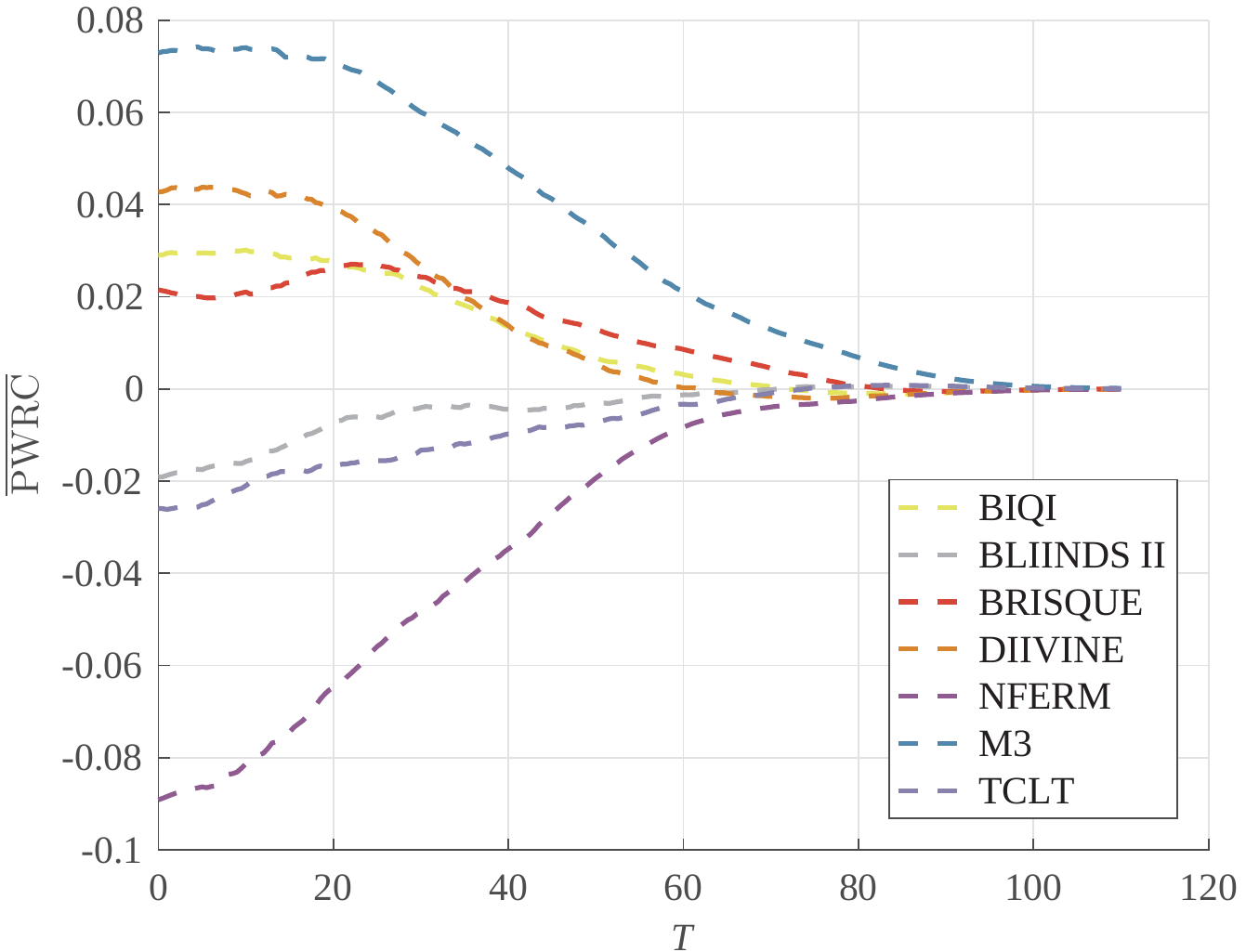}}
  \subfigure[ESPL-LIVE database with 80\% training data]{
  \includegraphics[width=.31\linewidth]{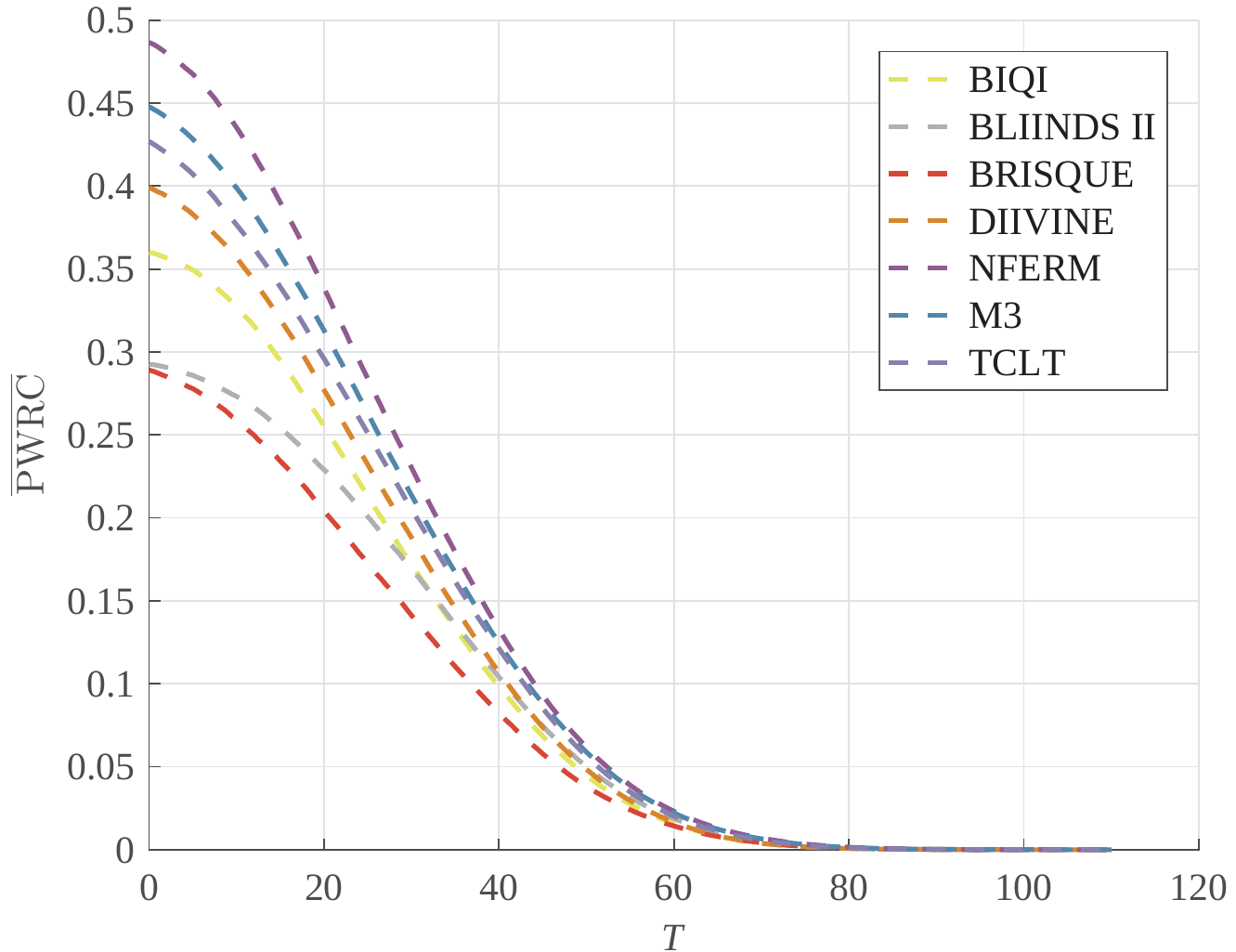}}
  \subfigure[ESPL-LIVE database with 50\% training data]{
  \includegraphics[width=.31\linewidth]{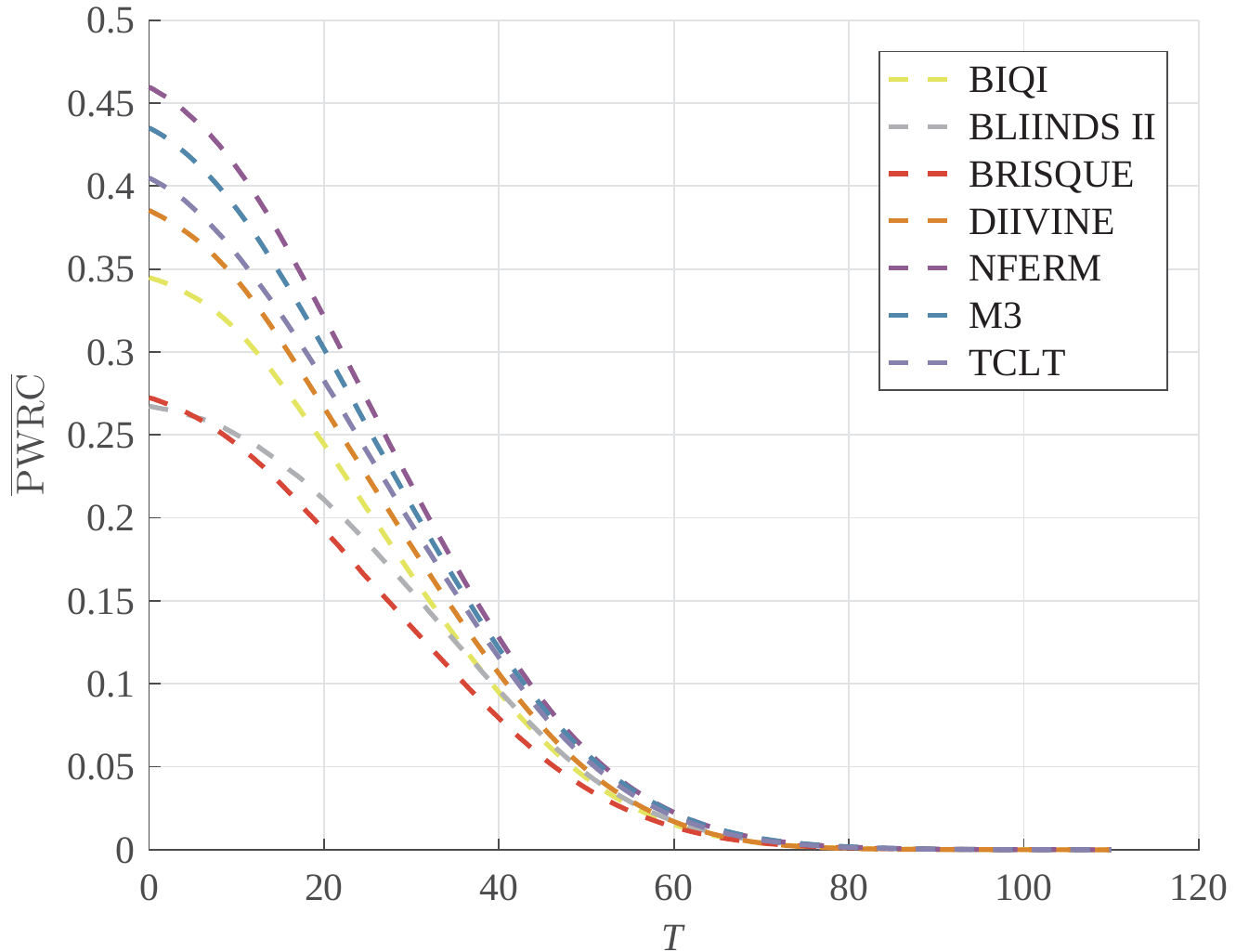}}
  \subfigure[ESPL-LIVE database with 20\% training data]{
  \includegraphics[width=.31\linewidth]{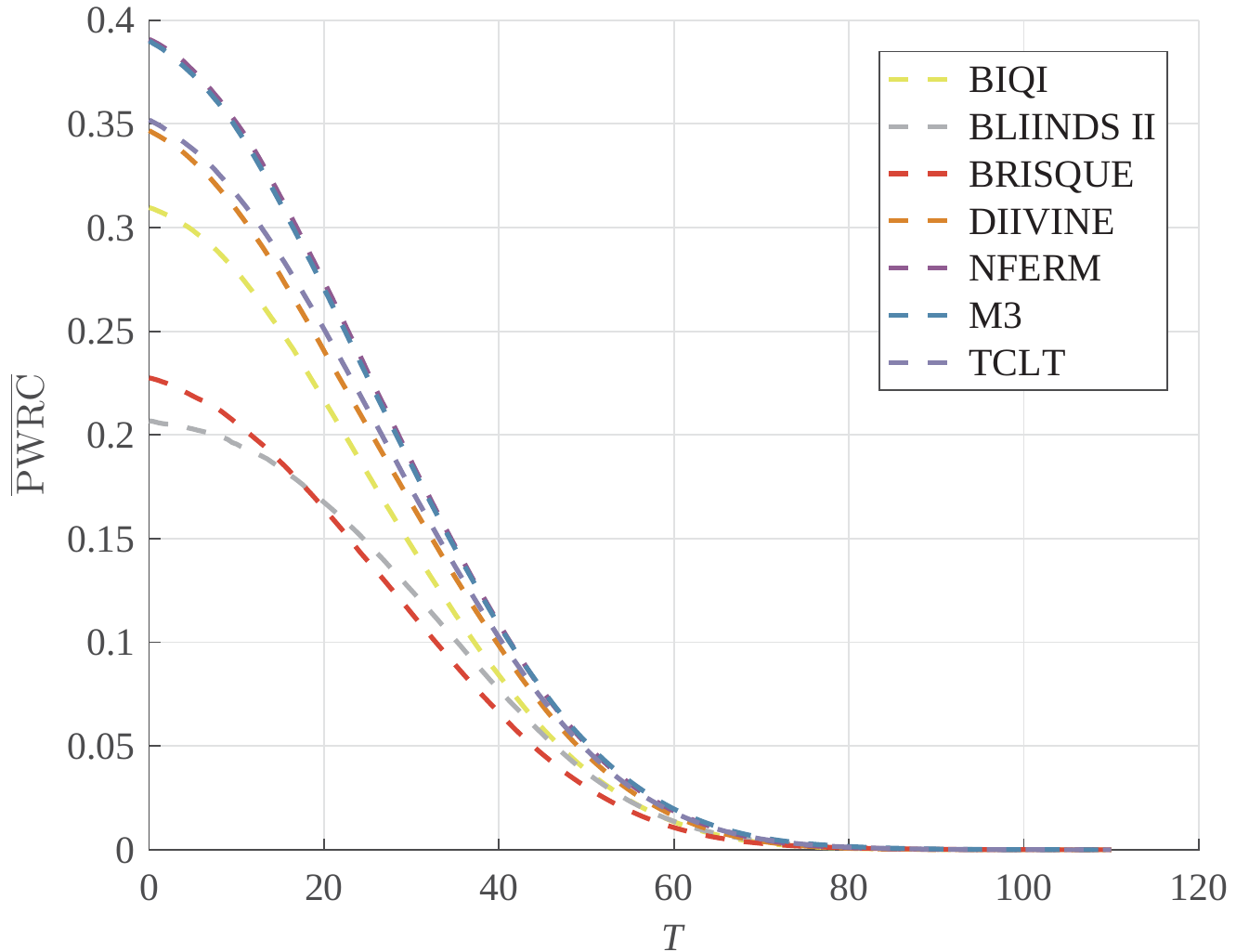}}
  \caption{The \textit{SA-ST} curves of different IQA algorithms tested on the enhanced images. (a)-(c) show the $\overline{\text{PWRC}}$ performances on the IVCDehazing database, where the training sets take up 80\%, 50\% and 20\% images, respectively. (d)-(f) give the $\overline{\text{PWRC}}$ results, whose training set ratios are 80\%, 50\% and 20\%, respectively.}
  \label{fig_enhancement_SAST_curves}
\end{figure*}

Reliable IQA metric plays a crucial role in kinds of quality driven applications. In this section, we utilize the proposed PWRC indicator to recommend appropriate IQA metrics for two image enhancement applications, i.e., image dehazing and HDR image reconstruction, which are investigated on IVCDehazing and ESPL-LIVE databases, respectively. More specifically, the IQA metric serves as a selector towards different enhancement algorithms. The PWRC aims to find the most robust selector from some candidate IQA metrics. It is worth nothing that there is no perfect reference for an enhanced image. So, we select seven state-of-the-art NR-IQA algorithms as the candidate metrics, which include BIQI \cite{BIQI}, BLIINDS-II \cite{BLIINDS-II}, BRISQUE \cite{BRISQUE}, DIIVINE \cite{DIIVINE}, NFERM \cite{NFERM}, M3 \cite{BIQA_GM_LOG} and TCLT \cite{TCLT}.

For fair comparison, different enhancement algorithms should be applied to the same visual content. In view of this fact, we develop a slightly deformed average MOS difference to represent the benchmark of each IQA metric in pushing perceptually preferred images. Given $K$ candidate algorithms and $L$ raw images, we use $\Delta \text{MOS}_{k,l}$ to denote the performance of the $k$th IQA metric for ranking $K$ enhanced versions of the $l$th raw image, where $\Delta \text{MOS}_{k,l}$ could be computed via Eq. (\ref{eq_ave_MOS}) by setting $n$ to $K$. Then, the overall performance of the $k$th IQA metric could be defined by
\begin{equation}
\Delta \overline{\text{MOS}}_k = \frac{1}{L}\sum_{l=1}^{L} \Delta \text{MOS}_{k,l}
\end{equation}
where the similar measurement could be found in the rational test of \cite{GMAD}.

Similar to $\Delta \overline{\text{MOS}}_k$, we use the image-wise averaged rank correlation indicators to sort different enhancement algorithms
\begin{equation}
\overline{\text{PWRC}}_k = \frac{1}{L}\sum_{l=1}^{L} \text{PWRC}_{k,l}
\label{eq_im_ave_MOS}
\end{equation}
where the area under the curve values computed from $\overline{\text{PWRC}}$ is denoted by $\overline{\text{AUC}}_{ca}$. For comparison, the deformed indicators $\overline{\text{SRCC}}$ and $\overline{\text{KRCC}}$ are also tested in this section, which can be computed by replacing $\text{PWRC}_{k,l}$ with $\text{SRCC}_{k,l}$ and $\text{KRCC}_{k,l}$ in Eq. (\ref{eq_im_ave_MOS}), respectively.

Similar to the test on degraded images, we first show the \textit{SA-ST} curves of different IQA metrics in Fig. \ref{fig_enhancement_SAST_curves}. It is seen that the existing NR-IQA metrics work poorly in recommending high quality dehazing or HDR image, whose $\overline{\text{PWRC}}$ values are all very small. In addition, in comparison with Fig. \ref{fig_SAST_curves}, we can clearly find more intersections between different \textit{SA-ST} curves as shown in Fig. \ref{fig_enhancement_SAST_curves}. For example, in Fig. \ref{fig_enhancement_SAST_curves} (a), the BRISQUE outperforms M3 when $T$ is smaller than 40. However, because the descent speed of BRISQUE is much higher than M3, its $\overline{\text{PWRC}}$ values become lower than M3 when $T$ is larger than 40. The similar observations could also be found between NFERM and BLIINDS II as shown in Fig. \ref{fig_enhancement_SAST_curves} (b). In addition, the intersection between BRISQUE and BLIINDS II also occurs in Figs. \ref{fig_enhancement_SAST_curves} (e) and (f), and so on. We can find that the intersection between different \textit{SA-ST} curves occurs more frequently for the IQA metrics whose $\overline{\text{PWRC}}$ locate in low values as shown in Figs. \ref{fig_enhancement_SAST_curves} (a)-(c). This is because that there are two factors could result in a low PWRC value, i.e., 1) a large number of correct ranks locate in low quality levels, or 2) a small amount of correct ranks locate in high quality levels. In case 1), the descent speed of \textit{SA-ST} curve would be relatively slow, where the changes in low quality levels only slightly impair $\overline{\text{PWRC}}$ due to their low weights. By contrast, in case 2), the descent speed of \textit{SA-ST} curve would be faster due to the larger weights assigned to high quality levels in $\overline{\text{PWRC}}$. This subtle discriminability enables $\overline{\text{PWRC}}$ to evaluate each IQA metric from a more comprehensive perspective. In the following, by means of $\overline{\text{AUC}}_{ca}$, we can derive a single-valued measurement from $\overline{\text{PWRC}}$ under the given confidence intervals.

In Tables \ref{table_Dehazing_results} and \ref{table_HDR_results}, we report the detailed accuracy of different IQA metrics on the IVCDehazing and ESPL-LIVE databases, respectively. Similar to the observations in \cite{IVCDehazing,HDR}, existing NR-IQA metrics work poorly in ranking the qualities of enhanced images. As shown in Table \ref{table_Dehazing_results}, many IQA metrics produce negative $\Delta \overline{\text{MOS}}$ values. The results in Table \ref{table_HDR_results} are slightly better, whose $\Delta \overline{\text{MOS}}$ values are still very small. All three rank correlation indicators could reflect the terrible accuracy of these algorithm selectors, where $\overline{\text{SRCC}}$, $\overline{\text{KRCC}}$ and $\overline{\text{AUC}}_{ca}$ are all very low. But, there are also disagreements between the detailed ranks of IQA metrics predicted by them, where the discordant IQA pairs are highlighted by boldface and italic in Tables \ref{table_Dehazing_results} and \ref{table_HDR_results}.

\begin{table*}[t]
  \centering
  \caption{Comparison of different rank correlation indicators on IVCDehazing database}
  \includegraphics[width = .8\linewidth]{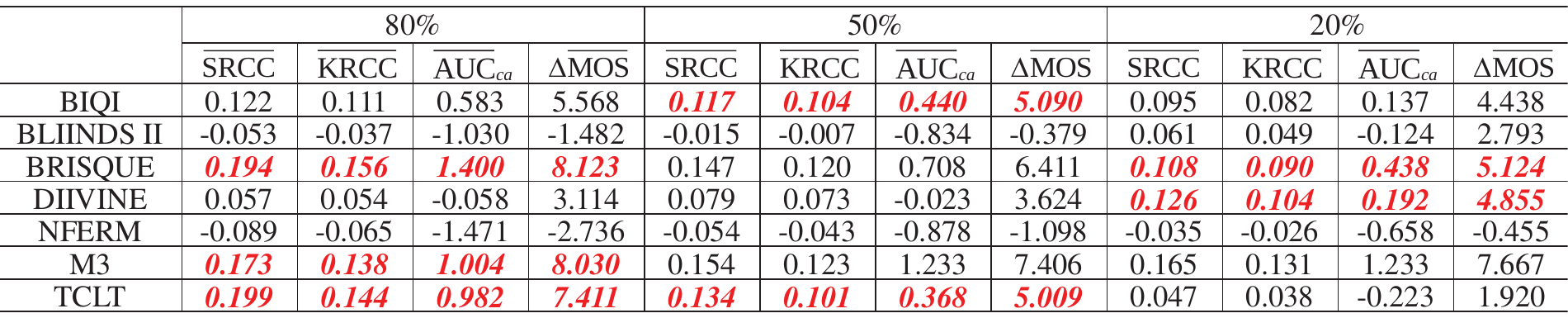}
  \label{table_Dehazing_results}
\end{table*}

\begin{table*}[t]
  \centering
  \caption{Comparison of different rank correlation indicators on ESPL-LIVE HDR database}
  \includegraphics[width = .8\linewidth]{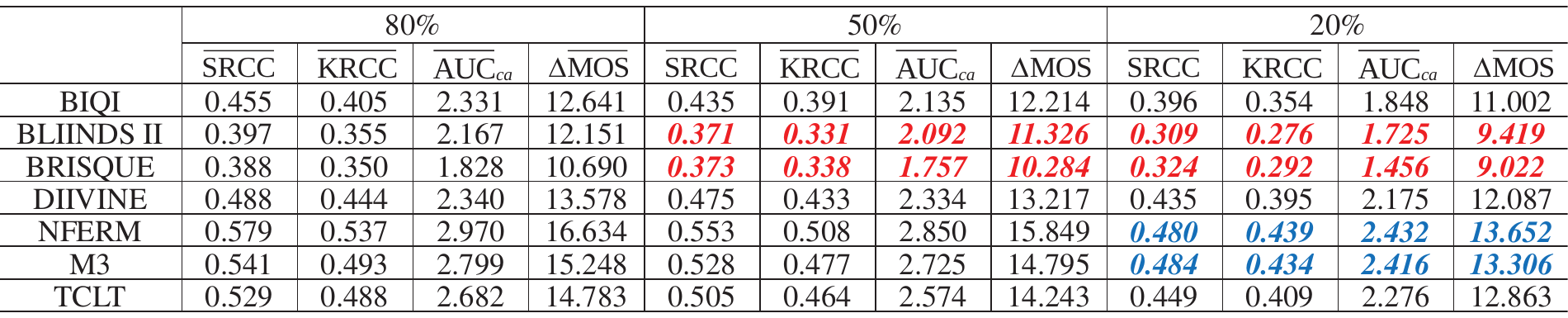}
  \label{table_HDR_results}
\end{table*}

As shown in Table \ref{table_Dehazing_results}, when 80\% dehazing images are used for training, both $\overline{\text{AUC}}_{ca}$ and $\overline{\text{KRCC}}$ recommend BRISQUE instead of M3 or TCLT, which is consistent with $\Delta \overline{\text{MOS}}$. While, the $\overline{\text{SRCC}}$ prefers TCLT.
In Table \ref{table_HDR_results}, when we use 50\% HDR images for training, both $\overline{\text{SRCC}}$ and $\overline{\text{KRCC}}$ consider BRISQUE is better than BLIINDS II. Only the proposed $\overline{\text{AUC}}_{ca}$ indicator is consistent with $\Delta \overline{\text{MOS}}$. The similar condition is also found for comparing BRISQUE with BLIINDS II, when the training set ratio is 20\%. These experiments confirm that the proposed $\overline{\text{PWRC}}$ indicator works better for recommending robust IQA metric towards enhanced images, whose top ranked samples would present higher average perceptual qualities.

\section{Conclusion}
In this paper, we propose a perceptually weighted rank correlation (PWRC) indicator to fairly compare different IQA algorithms. Inspired by two important visual perception properties, i.e., perceptual importance variation and subjective uncertainty, we develop the nonuniform weighting and adaptive activation schemes to evaluate the rank accuracy of each IQA algorithm. More specifically, a larger weight would be assigned to the image pair with higher quality level and greater rank deviation. Meanwhile, the comparison between two images would be activated only if their rank deviation exceeds the given sensory threshold, which suppresses the interference from perceptually unsortable image pairs. Extensive experiments on five publicly available IQA databases show that the proposed indicator is more consistent with human perception and works better for recommending perceptually preferred images.


%

\appendices
%

%
%

\ifCLASSOPTIONcaptionsoff
  \newpage
\fi



\bibliographystyle{IEEEtran}
\bibliography{UIRM}
\end{document}